\def\paperlanguage{} 
\pdfoutput=1 

\documentclass[letterpaper, 10 pt, conference]{ieeeconf}  

\usepackage{bm}
\usepackage{cite}
\usepackage{flushend}
\include{preamble}

\IEEEoverridecommandlockouts                              

\title{\LARGE \textbf
  {
    \switchlanguage%
    {%
      Development of Musculoskeletal Legs with Planar Interskeletal \\Structures to Realize Human Comparable Moving Function 
    }%
    {%
      面状骨格間構造を利用し任意姿勢でモーメントアームを確保し高出力での環境接触動作が可能な筋骨格脚の開発
    }%
  }
}

\author{Moritaka Onitsuka$^{1}$, Manabu Nishiura$^{2}$, Kento Kawaharazuka$^{1}$,\\Kei Tsuzuki$^{1}$, Yasunori Toshimitsu$^{1}$, Yusuke Omura$^{1}$,\\Yuki Asano$^{1}$, Kei Okada$^{1}$, Koji Kawasaki$^{3}$, and Masayuki Inaba$^{1}$
  \thanks{$^{1}$ The authors are with the Department of Mechano-Informatics, Graduate School of Information Science and Technology, The University of Tokyo, 7-3-1 Hongo, Bunkyo-ku, Tokyo, 113-8656, Japan.
    {\texttt\small [onitsuka, kawaharazuka, tsuzuki, toshimitsu, omura, asano, k-okada, inaba]@jsk.t.u-tokyo.ac.jp}
    }
  \thanks{$^{2}$ The author with the Graduate School of Interdisciplinary Information Studies at the University of Tokyo.
    {\texttt\small nishiura@jsk.t.u-tokyo.ac.jp}
  }
  \thanks{$^{3}$ The author is associated with TOYOTA MOTOR CORPORATION.
    {\texttt\small koji\_kawasaki@mail.toyota.co.jp}
  }
}
\begin{document}

\maketitle
\thispagestyle{empty}
\pagestyle{empty}

\begin{abstract}
  \switchlanguage%
  {%
Musculoskeletal humanoids have been developed by imitating humans and expected to perform natural and dynamic motions as well as humans. To achieve desired motions stably in current musculoskeletal humanoids is not easy because they cannot maintain the sufficient moment arm of muscles in various postures. In this research, we discuss planar structures that spread across joint structures such as ligament and planar muscles and the application of planar interskeletal structures to humanoid robots. Next, we develop MusashiOLegs, a musculoskeletal legs which has planar interskeletal structures and conducts several experiments to verify the importance of planar interskeletal structures. 
  }%
  {%
筋骨格ヒューマノイドは人体模倣を規範として開発され、人のように身体の柔軟性を活かした巧みな動作が期待される。一方で関節角度に応じて筋のモーメントアームが変化してしまうなど、筋骨格ヒューマノイドを様々な姿勢で安定して動作させるのは容易ではない。本研究では人のような動作を実現させるために関節を跨いで骨格に作用する筋や靭帯などの骨格間構造に見られる面状構造に着目し、その評価と実際のロボットへの適用について議論する。面状骨格間構造を有する筋骨格脚部を開発し、人と同様の関節の機能や、着席時の動作実現等を確認し、筋骨格ヒューマノイドにおける面状骨格間構造の重要性を示す。
  }%
\end{abstract}

\section{INTRODUCTION}\label{sec:introduction}
\switchlanguage%
{%
    The tendon-driven musculoskeletal humanoids \cite{SR2017:asano:design}, \cite{icra2006:kotaro} which imitates a human body structure, have muscles around bone structures as actuators. Muskuloskeletal humanoids are designed as a model of human beings and expected to perform natural and dynamic motions as a human being can do. Benefits of human mimetic structures are, for example, ball joints that have no singularity and redundant muscles, which enables variable stiffness control. 
However, muscle route can dynamically change in the current musculoskeletal humanoids as each muscle expands and contracts.
This change makes it challenging to maintain the sufficient moment arm of muscles in any posture.
Therefore it is difficult to apply enough torque to joints to perform the desired motions in some postures.
In this study, we focus on interskeletal structures, such as muscles and ligaments, which act over different skeletal structures to solve this problem.

Ligaments are the essential interskeletal structures which bond different skeletal structures like muscles, but they passively act with joints; on the other hand, muscles actively act. 
The primary function of ligaments is to make restrictions within joints and determine the relative direction of movements and the range of motion.
Even when the muscles do not perform tension, ligaments act passively to joints, stabilize joints and prevent the joint dislocation \cite{book:kapandji:legs}.
In many robots, axle-bearing and joint angle limits are implemented as a rigid mechanism.
These rigid joints restriction cannot handle impact force and often break irreversibly. 
Non-rigid hardware systems are required to realize joint structures that have an elastic and complex range of motion.

Previous researches on joint restriction using non-rigid materials are as follows, deformable membrane capsule for an open ball glenohumeral joint which restricts translational motion \cite{Biorob2018:fujii:capsule}, the ligament which restricts joints movement in rolling surface \cite{RoboSym:sonoda:ligaments} and anthropomorphic robotic hands that have silicon rubber sleeve ligaments \cite{ICRA2011:xu:hand}. 
The mechanism in \cite{Biorob2018:fujii:capsule} prevents dislocation by covering ball joints with rubber. 
This mechanism has a trade-off between the strength of rubber and the range of motion, and the deterioration of rubber can be inevitable. 
In \cite{RoboSym:sonoda:ligaments}, the bundle of fibers restricts joint movement in the knee to the rolling surface. 
This research only considers the two-dimensional movement, so this mechanism cannot handle the rotational movement of the knee.
The robotic hand in \cite{ICRA2011:xu:hand} has achieved flexible manipulation by using ligaments and shows the importance of ligaments in human-robotic hand structures.

To restrict spherical joints non-rigidly, wires and sleeves with high strength in the pulling direction are considered effective. 
However, using wires in three-dimensional structures is hard because wires cannot maintain their path under various configurations and wires tend to be caught in rough mechanical structures such as ditches and protuberance. 
In \cite{Humanoids2011:osada:planar}, the planar muscles solved some of these problems, but the insufficient durability in the shear direction is still yet to be solved.
}%
{%
筋骨格ヒューマノイド\cite{SR2017:asano:design}は人体を模した身体構造を特徴とし、骨格の周囲にアクチュエータとして筋を配置して駆動する人型ロボットである。人体模倣を規範として設計され、人の様な巧みな身体動作が期待される。人体の骨格に見られる柔軟な劣駆動多節構造を有する背骨や特異点の存在しない開放型球関節が実現されていることに加え、冗長に多数取り付けられた筋によって関節の可変剛性制御を可能としている。一方で関節角度によって筋のモーメントアームが変化して発揮可能なトルクが低下して、姿勢によって動作することが難しい場合がある。人のような動作の実現のためには、人の関節の構造や機能に学ぶ必要があり、本研究では骨格間を跨いで身体に作用する構造として筋や靭帯などの骨格間構造に着目する。

靭帯は筋と同様に関節をまたいで骨格同士を牽引する重要な構造だが、関節に対し受動的に働くという点で役割が異なる。靭帯の主な機能は関節に拘束を設け骨格間の相対運動の方向や可動域を決定することである。筋が脱力していたり有効に張力が伝えられない場合も受動的な力を関節に及ぼし、脱臼を防ぐなど関節の安定化に貢献している\cite{book:kapandji:legs}。多くのロボットでは軸受けや関節リミット機能をリジッドなハードウェアで実装しているが、リジッドな関節拘束は衝撃をハードウェアで吸収できず致命的な破壊が起きうる。人のような動作に耐え、柔らかで複雑な可動域を持つ関節を実現するためには靭帯を含めた関節構造を考える必要がある。

柔らかい素材による関節の拘束についての研究として開放型球関節の並進運動を拘束する関節包\cite{Biorob2018:fujii:capsule}や膝関節の運動を転がり曲面に拘束する靭帯ついての研究\cite{RoboSym:sonoda:ligaments}、人の手を精巧に模したロボットハンドで編地とゴムによって靭帯を実装した研究\cite{ICRA2011:xu:hand}がある。\cite{Biorob2018:fujii:capsule}は関節全体をゴムで包むことで人と同様の可動域を実現しつつ脱臼を防いでいるが、ゴムの強度と可動域にトレードオフがあり、広い可動域を確保するために強度を犠牲にしている上、ゴムの劣化も生じる。\cite{RoboSym:sonoda:ligaments}ではいくつかの繊維を束ねた靭帯によって人の膝における転がり運動への拘束を実現しているが2次元運動のみを扱っており、膝の回旋は考慮に入れていない。\cite{ICRA2011:xu:hand}では人体の関節構造を実装する際に靭帯を設けることで物体の柔軟な把持を可能とし、人体の手の構造における靭帯の有効性を示している。

人体に見られるような曲面形状を持った関節を柔らかに拘束するためには、引っ張り方向に強いワイヤや布等を用いた骨格間構造が適切だと考えられるが、3次元的に様々な方向への回転を許す場合、ワイヤによる骨格間の牽引は曲面上でワイヤがずれて骨格の支持ができない、機構の突起や隙間に引っかかりやすいなど実装上の問題が発生する。これは骨格に機構や組織をまとわせる筋骨格系に特有の問題である。\cite{Humanoids2011:osada:planar}では面状筋を実装することでこの問題を解決しているが、一方で骨格と線接触することによる機構への引っ掛かりを利用してモーメントアームを維持しているため人体の様に筋が骨格上を滑らず、ワイヤが平行に走行し滑車周りでワイヤが回転してしまうためせん断方向の力を発揮しづらいなど、課題が残されていた。
}%

\begin{figure}[t]
 \centering
  \includegraphics[width=.8\columnwidth]{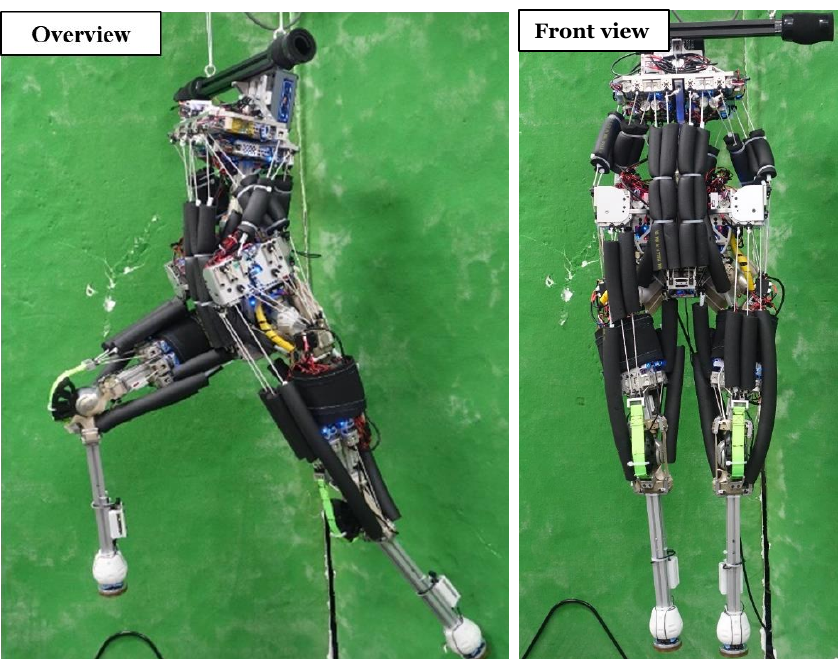}
  \vspace{-1.0ex}
  \caption{The musculoskeletal humanoid legs MusashiOLegs with planar interskeletal structures.}
  \label{fig:overview-of-musashiolegs}
  \vspace{-3.0ex}
\end{figure}

\switchlanguage%
{%
In this study, we focus on muscles and ligaments as interskeletal structures, implement human mimetic planar interskeletal structures in the musculoskeletal humanoid Musashiolegs (\figref{fig:overview-of-musashiolegs}), evaluate planar interskeletal structures and achieve a range of motion equivalent to human being and high torque performance in various postures. 
First, we describe the merits of planar interskeletal structures and evaluate them by modeling wires as nonlinear elastic units. 
Next, we discuss the application of planar interskeletal structures to musculoskeletal humanoids. 
Finally, we implement the musculoskeletal legs using planar interskeletal structures and confirm the robot has a function equivalent to a human being by conducting premilinary experiments.
}%
{%
そこで本研究では骨格間構造として筋と靭帯に着目し、人体における面状組織を模した構造、面状牽引構造を筋骨格ヒューマノイドに実装し、面状牽引構造の評価と人体の関節と同様の機能の実現を行う。
まず面状牽引構造の利点について述べ、ワイヤを非線形弾性要素とみなしてモデル化して評価を行う。次に面状骨格間構造の筋骨格ヒューマノイドへの適用について議論する。最後に筋骨格脚部の各関節に面状骨格間構造を実装し人と同様の機能が実現されていることを各種動作実験を通じて確認する。
}%

\section{PLANAR INTERSKELETAL STRUCTURE} \label{sec:musculoskeletal-structure}
\begin{figure}[htb]
 \centering
  \includegraphics[width=1.0\columnwidth]{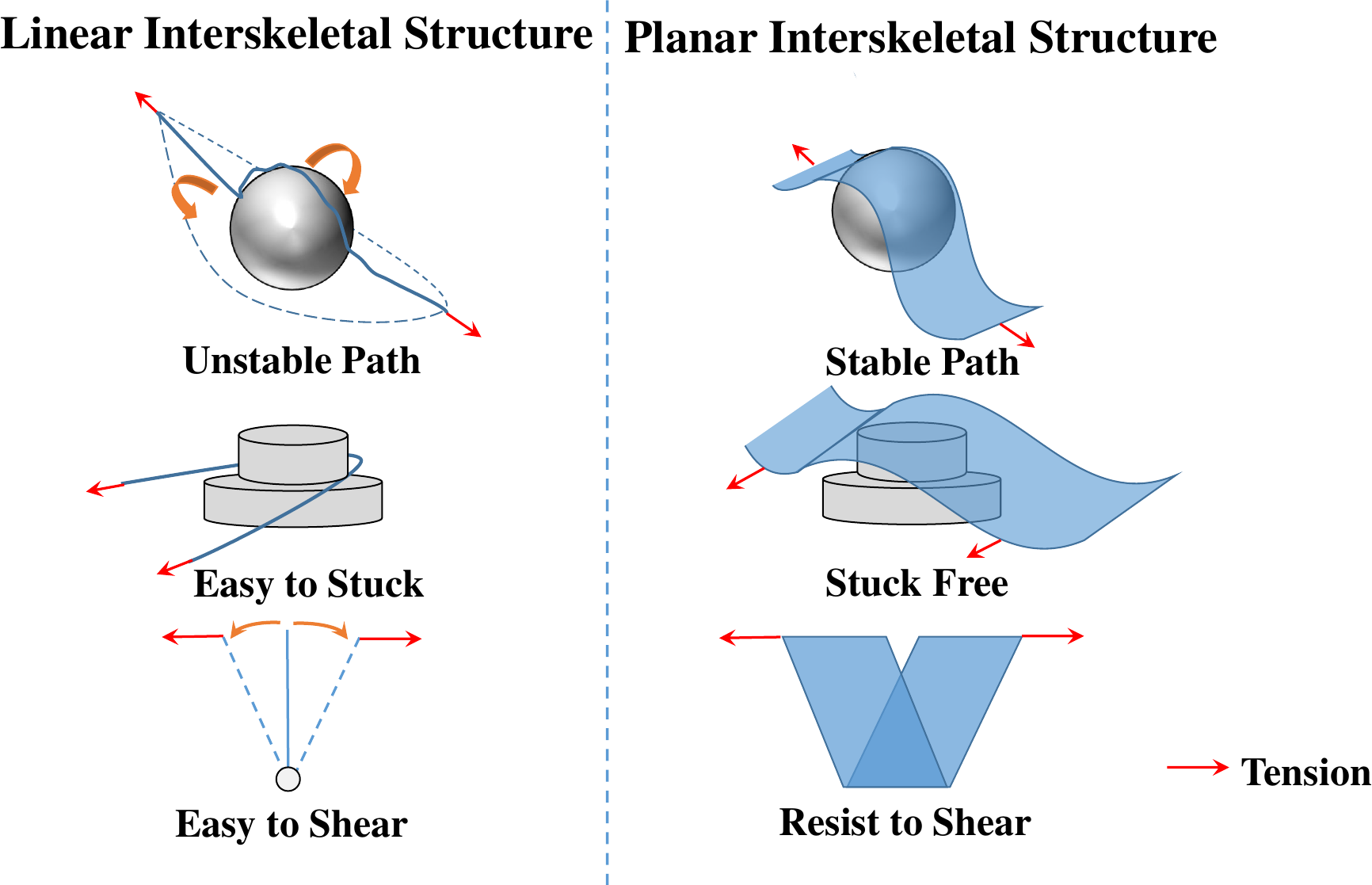}
  \vspace{-3.0ex}
  \caption{Comparison between linear interskeltal structure and planar interskeletal structure.}
  \label{fig:schema}
    \vspace{-3.0ex}
\end{figure}

\switchlanguage%
{%
    We define two types of interskeletal structures; linear interskeletal structure and planar interskeletal structure. 
Linear interskeletal structure is attached to skeletal structures with points and linearly contacts with skeletal structures. 
Planar interskeletal structure is attached to skeletal structures with lines and planarly contacts with skeletal structures. 
The planar interskeletal structure has the following three merits: (1) it can maintain the stable path; (2) it has stuck-free design; (3) it has high durability to the shear force.
}%
{%
筋や靭帯などの骨格間構造について、ワイヤや紡錘形のように骨格に点で取り付けられ線状に接触するものを線状骨格間構造とし、骨格に線で取り付けられ面で接触する構造を面状骨格間構造とする。\figref{fig:schema}に線状骨格間構造と面状骨格間構造の比較の模式図を示す。面状骨格間構造は線状骨格間構造に比べて次の利点を持つ。
}%

\subsection{Stable Contact Realized by Planar Structures}\label{subsec:stable-contact}
\switchlanguage%
{%
When the planar interskeletal structure contacts with the spherical skeletal structure, it can fit the skeletal structure by changing its path to transit to a lower energy potential state.
This characteristics of the planar interskeletal structure keeps the skeletal structure inside of its surface. 
However, the skeletal structure under the linear interskeletal structures can easily deviate from them. 
When the joint deviates from the linear interskeletal structure, it is hard to support or control the joint by ligaments or muscles \cite{Humanoids2011:osada:planar}. 
On the other hand, the planar interskeletal structure can stably transmit the muscle tension to the joint because it can maintain the planar contact with the skeletal structure even if the joint angle changes.
}%
{%
    球体関節などの曲面形状を持った骨格に骨格間構造が接する時、骨格間構造の両側には張力がかかっている場合力学的なポテンシャルが低い経路を取るように形状に変形していく。面状骨格間構造の場合、面接触によって骨格が面の内側から外側に逸脱することは無いが、線状骨格間構造の場合容易に逸脱する。ワイヤを用いた靭帯や筋の場合、関節角度に応じて関節が骨格間構造の外側に出てしまい、関節の支持や制御が不可能になる\cite{Humanoids2011:osada:planar}。面状骨格間構造は骨格と面接触を保つことで、関節角度が変化しても安定してその張力を関節に伝えることができる。
}%

\subsection{Catching Prevention Mechanism by Planar Structure}\label{subsec:catching-prevention}
\switchlanguage%
{%
The linear interskeletal structure has small contact areas and is often caught in roughness of skeletal structures. 
This phenomenon increases involuntary tension between tendons in linear interskeletal structures, and therefore forcibly causes the change of muscle path and increases friction. 
Compared to linear interskeletal structures, planar interskeletal structures have large contact surfaces with skeletal structures and cover skeletal structures by their surfaces. 
This feature prevents the planar interskeletal structure from being caught in the folds of skeletal structures.
}%
{%
    線状骨格間構造は骨格との接触面積が小さく、機構上生じてしまう溝や角に引っかかりやすい。これによって意図しない張力が生じ、筋の意図しない経路変化や摩擦による動作不良などの問題が生じていた。
面状骨格間構造は骨格形状を覆うように変形するため、溝や角に引っかかりにくい。
}%

\subsection{Durability to Shear Force}\label{subsec:resistance-shear}
\begin{figure}[htb]
 \centering
 \includegraphics[width=1.0\columnwidth]{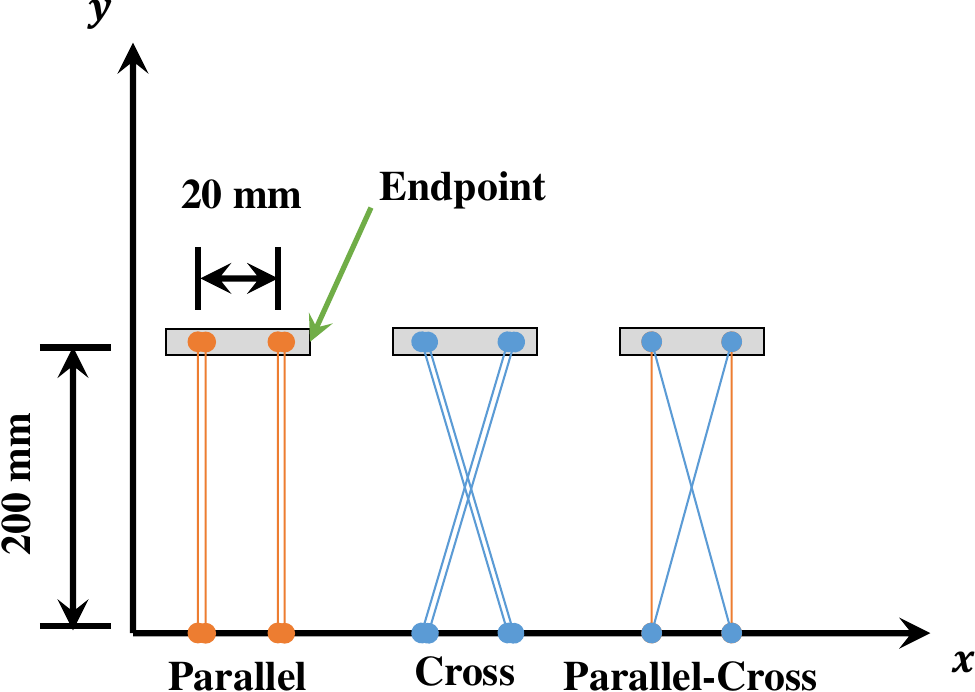}
  \vspace{-3.0ex}
  \caption{Schematic diagram of traction structures. The planar structure is simplified as wires which pulls parallel and cross direction}
  \label{fig:comparison}
    \vspace{-1.0ex}
\end{figure}

\switchlanguage%
{%
The linear skeletal structure can handle force in the pulling direction by its elasticity on the condition of being fixed by their position (rotationally free), but can hardly handle force in the orthogonal direction. 
Therefore the planar muscle \cite{Humanoids2011:osada:planar}, which is simply constructed by parallel muscles, cannot produce high power in the shear direction. 
On the other hand, we expect the planar interskeletal structures to handle force in the shear direction by interfering with fibers. 
We model the elasticity of planar interskeletal structures as a cluster of fibers and compares it with some liner interskeletal structures. 
\figref{fig:comparison} shows schematic diagram of skeletal structures. 
}%
{%
端点が位置のみ拘束を受け、回転について自由度を持つとすると、線状骨格間構造は引っ張り方向にはその弾性によって力を受けることができるが、それと直交する方向には力を受けづらい。したがってワイヤを平行に並べた従来の面状筋\cite{Humanoids2011:osada:planar}ではせん断方向に大きな力は発揮できない。一方面状骨格間構造は繊維同士が干渉しあうことでせん断方向の力も受けることができると考えられる。面状骨格間構造の弾性をワイヤの集まりとしてモデル化し、線状骨格間構造の複数の牽引方法と比較して評価する。\figref{fig:comparison}に線状骨格間構造と面状骨格間構造の比較の模式図を示す。
}%

\begin{figure}[htb]
 \centering
  \includegraphics[width=1.0\columnwidth]{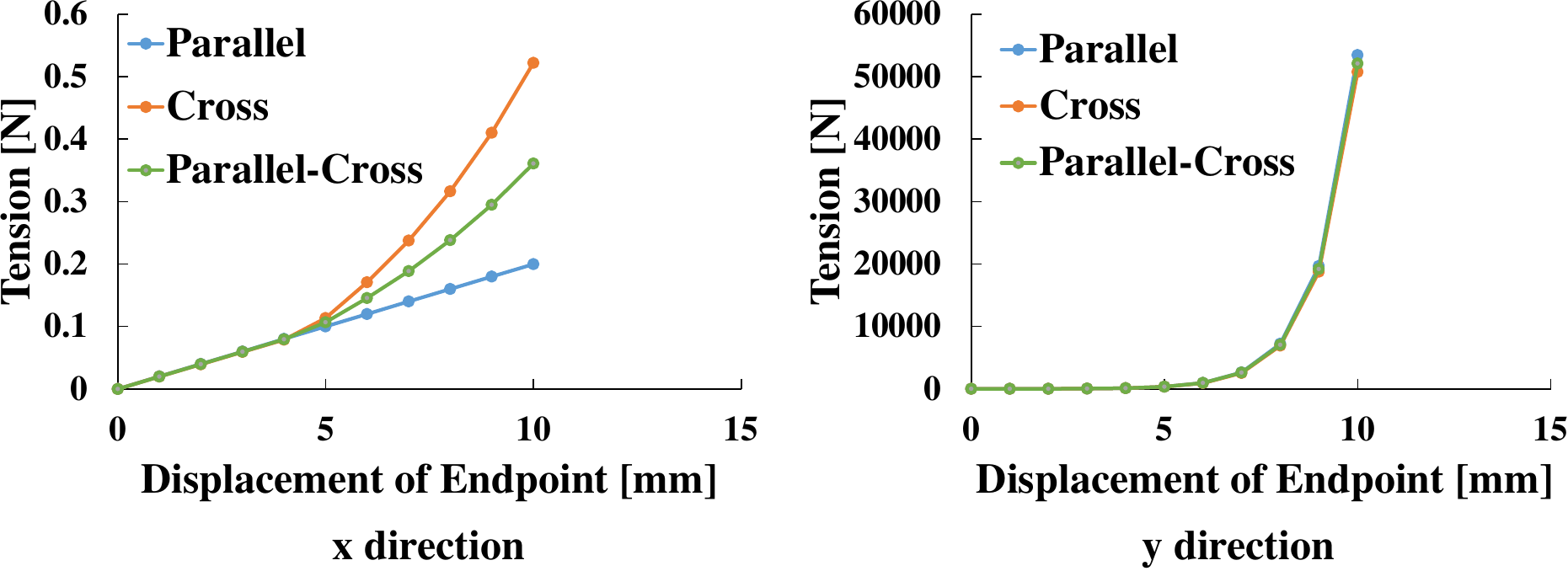}
  \vspace{-3.0ex}
  \caption{Simulation result of relationship between wire tension and displacement of endpoint. The endpoint is tracted in x and y direction independently.}
  \label{fig:simulation}
  \vspace{-1.0ex}
\end{figure}

\begin{figure}[htb]
 \centering
  \includegraphics[width=0.9\columnwidth]{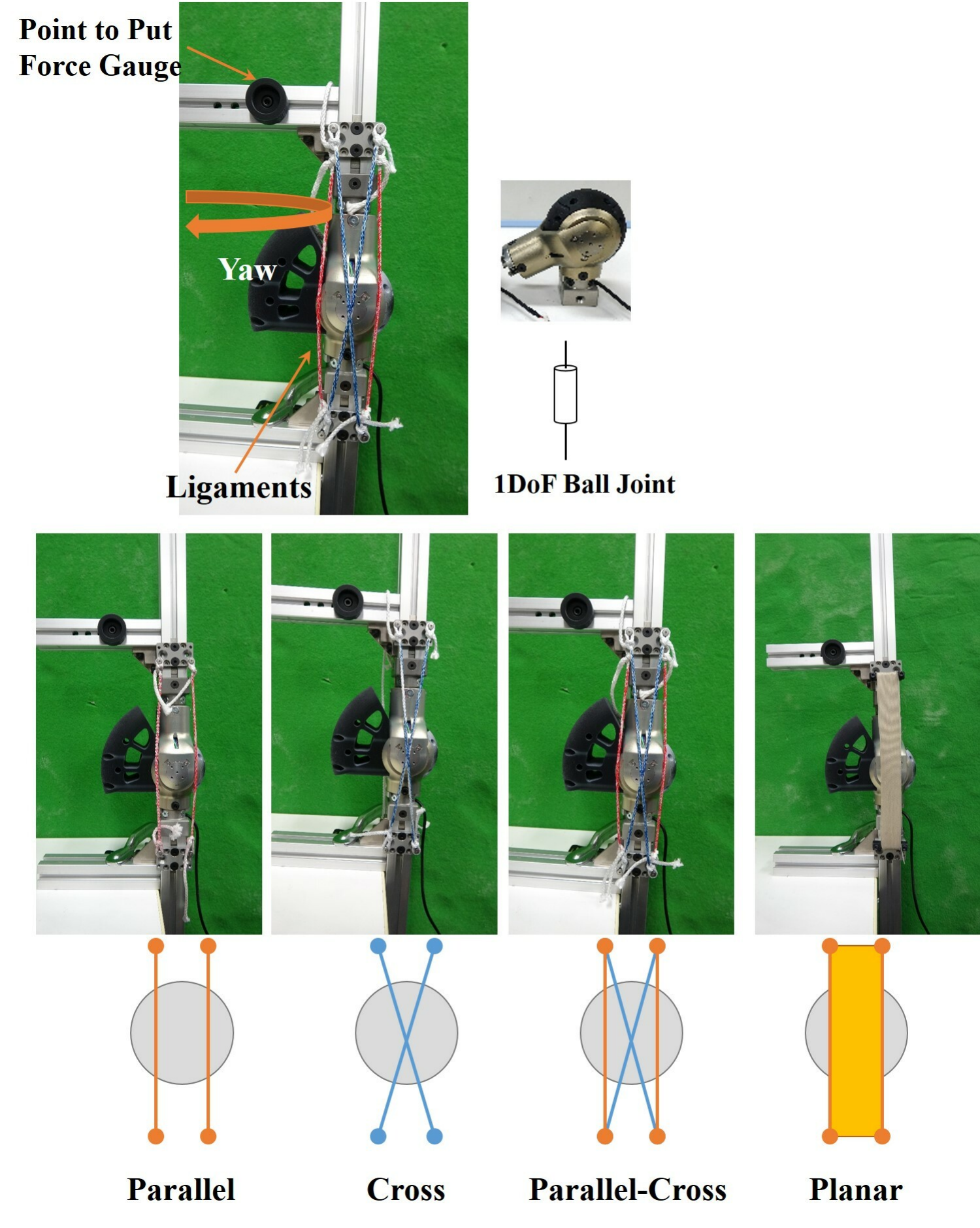}
  \caption{Experimental settings for the actual verification of the relationship between the applied torque and the yaw angle of the knee joint. In addition to the three wire settings, the planar structure made of cloth was also verified.}
  \label{fig:experimental-settings}
\end{figure}

\begin{figure}[htb]
 \centering
  \includegraphics[width=0.9\columnwidth]{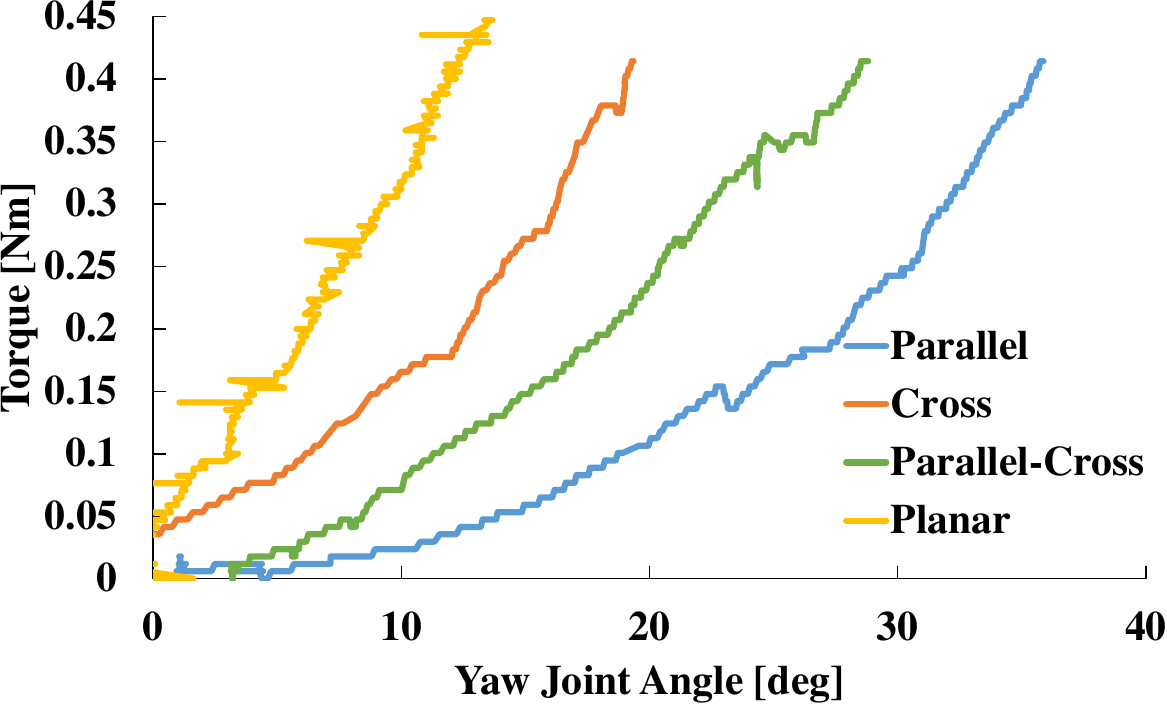}
  \caption{Actual verification result of the relationship between the applied torque and the yaw angle of the knee joint. The result shows similar tendency to the theoretical result.}
  \label{fig:passive-compliance}
\end{figure}

\switchlanguage%
{%
    Breen, et al. modeled the precise model of woven cloth based on the minimization of energy \cite{Breen:1994}. However, in this study, we model woven cloth as truss structure which is tracked by parallel and crossed wires (as shown in parallel-cross in \figref{fig:comparison}), to consider external forces in only two directions on surface. The model which is tracked by parallel wires (\figref{fig:comparison} parallel) and the model which is pulled by crossed wires (\figref{fig:comparison} cross) will be compared with the planar-coss model. We assume that the tension and the elongation of wires have the following relationship.
\begin{equation}
  T = \exp(K\max(0,(l-l_0) - l_d))
\end{equation}
${T}$ is the tension and ${K}$ is the stiffness of wire. ${K}$ is multiplied by the difference of current length of wire ${l}$ and equilibrium length of the wire ${l_0}$. 
We modeled it so that the tension of the wire develops according to the exponential function as the wire elongates. 
The tension develops only when the wire elogates over the constant elongation. 
The dead-band of the each wire is setted in $(-\infty,l_d]$.
We set ${K=1.0}$/mm, ${l_d=0.5}$mm. ${l_0}$ is calculated based on the assumption that mounting points are 200mm away in the direction of ${y}$ axis in \figref{fig:comparison}, 20 mm away in ${x}$ axis. 
On the condition that each structure has 4 wires, the result is shown in \figref{fig:simulation}. 
It shows that parallel-cross structure can bear the force of the shear direction more than parallel structure and cross structure. 
In addition, this result shows that three sutructures have almost same durability in the pulling direction, but parallel-cross structure is a little stronger than cross structure in the pulling direction. 
Parallel structure is most strongest in the pulling direction.
However, in order to use the advantage of merits described in \ref{subsec:stable-contact} and \ref{subsec:catching-prevention}, planar structure which is modeled as parallel-cross sturucture is necessary, because linear sturcture cannot use the advantage of these merits.

The actual elasticity of each model against shear force was verified in an experiment using wires by Dyneema.
An overview of the experiment is shown in \ref{fig:experimental-settings}.
In this experiment, the wires were attached in such a way that restoring force was generated for the yaw degree of freedom of the knee joint as described in \ref{sec:human-comparable-SHM}.
In addition to the three wire attachment settings, a planar structure made of cloth was added for comparison.
The results of the experiment are shown in \ref{fig:passive-compliance}.
The result shows similar tendency to the simulation result. 
Although it is not possible to simply compare wires and a cloth because of the different materials and attachment methods, the planar structure made of a cloth is superior to the wire in terms of elasticity.
}%
{%
布の形状のモデル化としてはエネルギー最小化に基づいてBreen \cite{Breen:1994}らによって正確なモデルが得られているが、本研究では面上の二方向の外力に対する弾性のみを考えるため、布のモデルとして\figref{fig:comparison}のParallel-Crossにあるとおり、並行なワイヤと交差するワイヤによって牽引されるトラス構造で簡略化して考える。また線状骨格間構造との比較として並行なワイヤのみで牽引した場合(\figref{fig:comparison} Parallel)と交差したワイヤのみで牽引した場合(\figref{fig:comparison} Cross)を考える。ワイヤの張力と伸びの間に次の関係が成り立っているとする。
\begin{equation}
  T = \exp(K\max(0,(l-l_0) - l_d))
\end{equation}
${T}$は張力、${K}$はワイヤの剛性を示し現在のワイヤの長さ${l}$と自然長${l_0}$の差分にかかる。ワイヤの伸びについて$(-\infty,l_d]$の範囲に不感帯を設け、一定以上の伸びに対して非線形かつ急激に張力を増大させるワイヤを仮定して指数関数として表す。
${K=1.0}$/mm、${l_d=0.5}$mmとし、${l_0}$は取り付け位置の間隔が\figref{fig:comparison}の${y}$方向に200mm、${x}$方向に20mm離れているとして\figref{fig:comparison}の各構造についてワイヤの張力と端点の変位の関係を計算した。ワイヤの本数をそれぞれ四本とし、その結果を\figref{fig:simulation}に示す。これによってワイヤを平行に並べた構造よりも互いの繊維が干渉する構造の方がせん断力を受けることが出来ると分かる。また交差させた構造に対してトラス構造の方が僅かに引っ張り方向の力を受けることができていることが分かる。

せん断力に対する各モデルの弾性を、Dyneemaによるワイヤを用いて実際に検証した結果を\figref{fig:passive}に示す。この検証では後述の膝関節のyaw自由度に対し復元力が生じるようにワイヤを取り付けた他、3つのワイヤモデルに加えて実際に布によるPlanar構造を比較対象に加えた。理論的な結果と同様の結果が得られていることが分かる。素材が異なることに加え取り付け方法も異なるためワイヤと布を単純に比較することは出来ないが、布が弾性においてワイヤに勝るという結果も得られた。
}%

\section{DEVELOPMENT OF MUSCULOSKELETAL LEGS WITH  PLANAR INTERSKELETAL STRUCTURES} \label{sec:application}
\switchlanguage%
{%
    In this section, examples of applying the interskeletal planar structure to a humanoid robot will be described. 
First, the specification of ''MusashiOLegs'' will be described. 
Next, the requirements for the application of the planar interskeletal structure to musculoskeletal legs will be described. 
Finally, we will describe the implementation of the planar interskeletal structure to the MusashiOLegs.
}%
{%

続いて面状骨格間構造の筋骨格ヒューマノイドへの適用について述べる。まず本研究で開発した筋骨格下肢である``MusashiOLegs''の仕様について述べる。次に筋骨格ヒューマノイドへ面状骨格間構造を適用する際の要求仕様について述べ、最後にMusashiOLegsへの実際の面状骨格間構造の実装について紹介する。
}%

\subsection{Musculoskeletal Humanoid ``MusashiOLegs''}
\begin{figure}[htb]
 \centering
  \includegraphics[width=1.0\columnwidth]{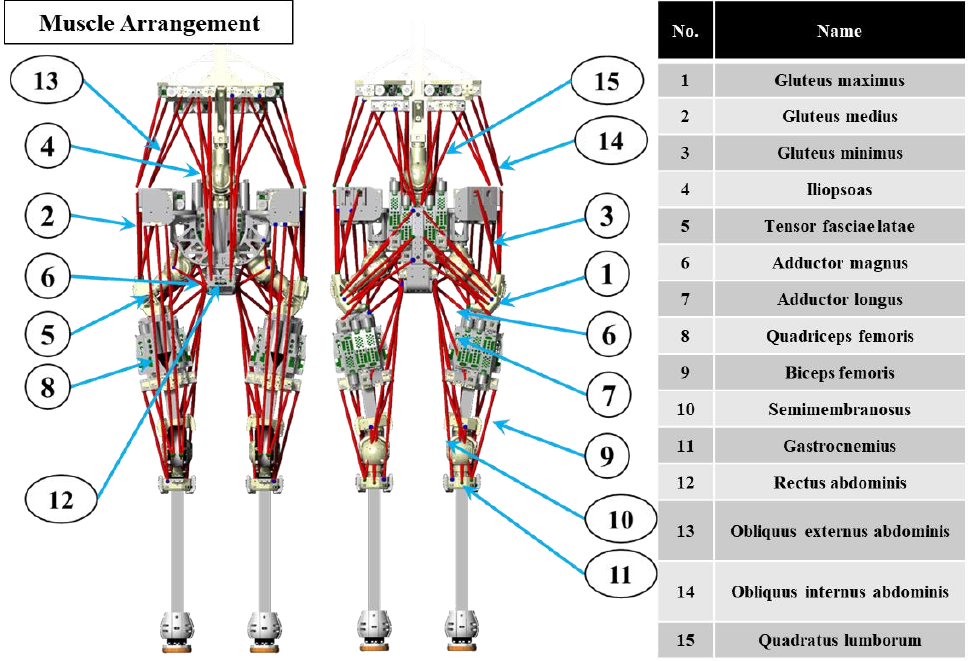}
  \caption{Overview and muscle arrangement of the musculoskeletal humanoid MusashiOLegs.}
  \label{fig:MusashiOLegs}
    \vspace{-3.0ex}
\end{figure}

\switchlanguage%
{%
    In this research, we developed the musculoskeletal humanoid legs MusashiOLegs (\figref{fig:MusashiOLegs}) as a successor of the musculoskeletal humanoid Musashi \cite{IROS2019:kawaharazuka:musashi}. 
MusashiOLegs has 3 DOF in the spine joint, 3 DOF in the hip joints and 2 DOF in the knee joints.
Each joint are constructed with pseudo spherical joints \cite{IROS2019:kawaharazuka:musashi}. 
We show the overview and the muscle arrangement of MusashiOLegs in \figref{fig:MusashiOLegs}.
The skeletal structures are drived by the contraction of the muscles (BLDC motors wind the chemical fiber Dyneema) \cite{asanoSensordriverIntegratedMuscle2015}. 
It has 40 actuators as muscles, and some wires are folded back to gain enough torque with relatively few actuators.
}%
{%
本研究では面状骨格間構造を有する筋骨格ヒューマノイドとして開発された、Musashi\cite{IROS2019:kawaharazuka:musashi}シリーズの新型機となるMusashiOLegs(\figref{fig:MusashiOLegs})を用いる。MusashiOLegsは脊椎関節に3自由度、股関節に3自由度、膝関節に2自由度の全13自由度を有し、各関節はKawaharazukaらによる擬似球関節を用いている。筋骨格系は\figref{fig:MusashiOLegs}の右図の様になっており、BLDCモータによって化学繊維であるダイニーマによるワイヤを巻き取り、筋の収縮を再現して骨格を駆動する腱駆動方式をとっている。筋として全身で40本のアクチュエータを使用しており、比較的少ないアクチュエータで十分なトルクを得るため、ワイヤの折り返しや多関節筋を多用している。
}%

\subsection{Design Requirements of Planar Interskeletal Structures}
\switchlanguage%
{%
    This section discusses the requirements of applying the interskeletal planar structure to humanoid robots. 
We classify the interskeletal structures into two types; the passive planar interskeletal structure and the active planar interskeletal structure. 
The passive planar structure cannot produce force by itself and only transmit force to the other body parts like ligaments. 
The active planar structure exists in actively drivable body parts such as muscles.
}%
{%
本節では面状骨格間構造を筋骨格ヒューマノイドに取り付けていく際の要求仕様について述べる。
}%

\subsubsection{Passive Planar Interskeletal Structures} \label{subsubsec:passive-pis}
\begin{figure}[htb]
  \centering
  \includegraphics[width=1.0\columnwidth]{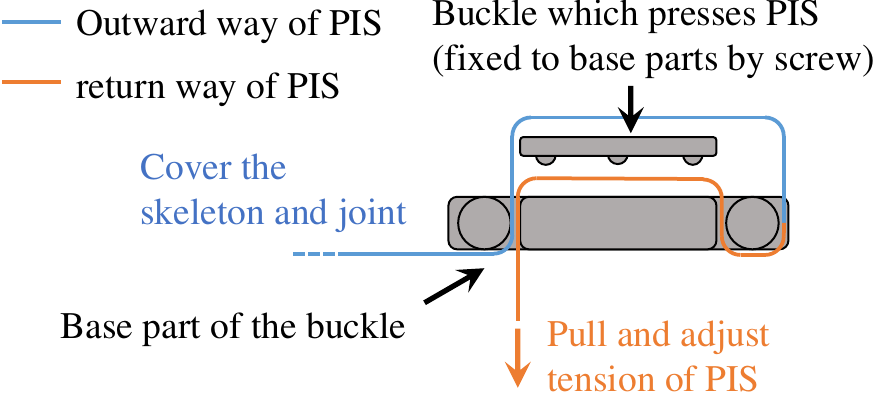}
  \vspace{-3.0ex}
  \caption{Schematic diagram of buckle mechanism. This mechanism can adjust the tension of the planar inter skeletal structure(PIS).}
  \label{fig:buckle}
\end{figure}

\switchlanguage%
{%
    Ligaments is an example of interskeletal structures which act passively to joints.
The main role of passive planar interskeletal structures is to restrict the range of joint angles and stabilize joint movement.
These inskeletal structures cannot produce tension or torque by itself.
Therefore it is necessary to attach skeletal structure by human so that it can produce tension only at the angle limit.
Next, we clarify the requirements to design a passive planar interskeletal structure as a body part of a musculoskeletal humanoid.

We use non-rigid material to achieve soft angle limit like a human being.
Continuously applying tension to non-rigid material lead to irreversible change of its sturcture and weaken the tension generated by its material.
To keep constant tension at the angle limit, the tension of elastic material should be adjusted if it is necessary.
Makino, et al. developed a mechanism which can restrict DOF in finger joints using linear interskeletal structures (wires) in a five-fingered hand with machined springs \cite{IROS2018:makino:hand}. 
In \cite{IROS2018:makino:hand}, they have shown twisting wires can adjust the equilibrium length and the tension of the ligament. 
However, when using the planar interskeletal structure to restrict joints, we cannot twist the planar structures, such as woven cloth.
The planar structure should be fixed to skeletal structure with line contact so that the planar structure can be pulled as 2d structure.
In addition, to prevent the interference of the interskeletal structure with muscles, the attachment part should be compact.

To summarize these conditions, the requirements are the following.
\begin{itemize}

\item Easy to adjust the tension

\item Capable of fixing the interskeletal structure to the skeletal structure by line contacts

\item Compact structure

\end{itemize}

We adapt buckle mechanism (\figref{fig:buckle}) to satisfy these requirements. 
This mechanism can strongly hold ligament by pressing force and friction of cover parts and can easily change the length of the ligament. 
In addition, separating this mechanism into two parts makes this mechanism compact.
}%
{%
靭帯など、関節に受動的に作用する人体の組織を面状骨格間構造として筋骨格ヒューマノイドに適用する場合の要件について考える。
本研究では筋骨格ヒューマノイドの関節として用いる関節モジュールがすでに関節の回転自由度を拘束しているため、関節可動域の決定や安定化が主な役割となる。
このような骨格間構造は自ら駆動することが出来ないため、拘束したい可動域限界において張力が発揮するように骨格間構造を固定する必要がある。

一方柔軟な素材での拘束を考慮すると、繊維の不可逆な伸びが発生することから固定時の張力を調整できる取付方法をとる必要がある。
固定時の張力を調整できる取り付け方法が望ましい．
Makinoらは切削ばねによる五指ハンドにおいて、線状骨格間構造であるワイヤによる靭帯によって指関節の自由度を拘束している\cite{IROS2018:makino:hand}。
この場合ワイヤを捻ることによって固定時の靭帯の自然長や張力を調整しているが、面状骨格間構造で関節を拘束するためには靭帯となる布などの面構造を捻るわけには行かないため、別の固定方法を考える必要がある。
面状の弾性体が面状構造として機能するためには骨格構造に対して線でとりつける必要がある．
また骨格に取り付けられ骨格に巻き付いて動く筋との干渉を抑えるため、取り付け部はコンパクトであると良い。

以上から取り付け部の設計要件をまとめると以下のようになる
\begin{itemize}

\item 張力調節が用意

\item 骨格間構造を骨格に対して線で固定可能

\item コンパクトな構造

\end{itemize}

以上の要件が達成可能な固定方法としてバックル方式(\figref{fig:buckle})による固定方法を考案した。バックル方式はねじによる押付力とベルトとなる靭帯による押付力によって強固に靭帯を保持できる上に靭帯の長さの調整も容易だが、部品点数が大幅に増え、靭帯が折り返すスペースが必要なことからコンパクトさは犠牲となった。
}%

\subsubsection{Active Planar Interskeletal Structures} \label{subsubsec:active-pis}
\switchlanguage%
{%
In comparison to a ligament, the muscle can actively contract. 
Ligaments are required to perform tension around the angle limit of the joint. 
Planar muscles are on the contrary required to perform and transmit torque generated from each muscle in any configuration. 
When planar muscle is actuated, muscles contact strongly with skeletal structures and there can be large friction.
Therefore we need to reduce friction between muscles and skeletal structures in order to keep muscles moving.
To summarize these conditions, the requirements which can realize the actively drivable planar interskeletal structure are following.

\begin{itemize}
    \item Structure which can maintain the sufficient moment arm to perform high joint torque
    \item Surface material which enables smooth contact of muscle with a skeletal structure
\end{itemize}

The smooth contact of muscles with skeletal structure is considered to be especially important. 
In the human body, a sheath of tendon and a fascia plays this role.

}%
{%
筋は先述の靭帯と異なり能動的に伸縮が可能であるため、腱駆動方式を想定する本研究では前節で述べたように張力を維持して固定する必要性は強くない。
一方関節可動域の限界において緊張していればよかった靭帯に対し、可動域中の様々な関節位置にあっても筋張力から生じるトルクを発生させることが求められる。
駆動時に筋は常に緊張し骨格と強く摩擦することが考えられる。
したがって広い可動域を持つ関節において能動的に駆動可能な骨格間構造に求められるのは以下の通りとなる。

\begin{itemize}
\item 大きなトルクを発生させるためのモーメントアーム維持可能な構造
\item 骨格とのなめらかな接触を実現する表面素材
\end{itemize}

特に骨格との滑らかな接触は、人体において腱鞘や筋膜が担っているように、人体模倣としても重要だと考えられる。
}%

\subsection{Application of Planar Interskeletal Structures to Musculoskeletal Legs ''MusashiOLegs''}
In this subsection, we discuss the application of planar interskeletal sturcture to the musculoskeletal humanoid. 
First, we describe the two passive planar interskeletal structures; iliofemoral ligaments and knee collateral ligaments. 
Next, two active planar interskeletal structures is described; pattella ligaments and gluteus maximus.
\subsubsection{Iliofemoral Ligaments}
\begin{figure}[htb]
  \centering
  \includegraphics[width=1.0\columnwidth]{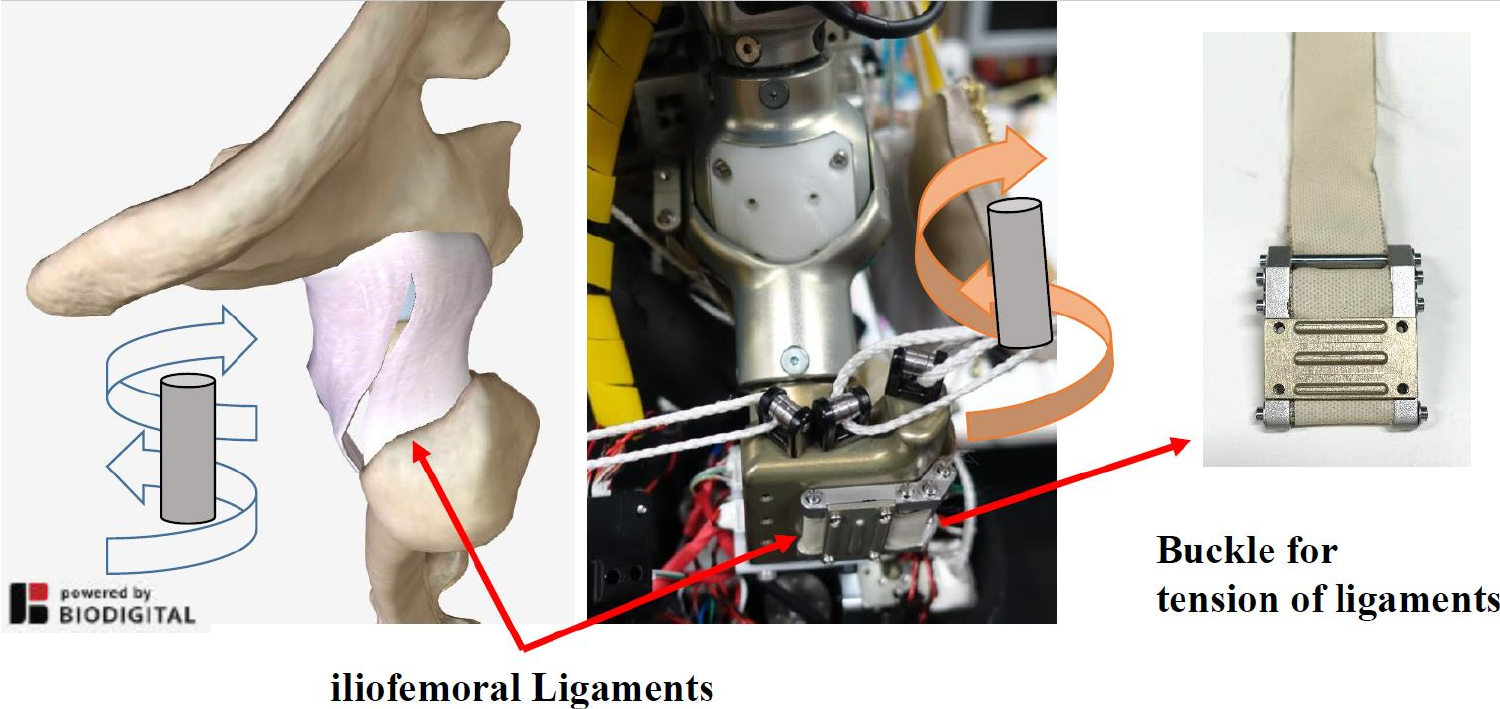}
  \caption{3D musculoskeletal model of the hip joint \cite{biodigital} and the application of iliofemoral ligament as planar interskeletal structure. Iliofemoral ligament is tensioned by the buckle.}
  \label{fig:application-hip-spiral}
\end{figure}

\switchlanguage%
{%
    The iliofemoral ligament is the ligament that winds around a caput femoris. 
The main role of this ligament is to restrict the hip joint angles. 
The iliofemoral ligament relieves when the hip joints are bent. 
On the other hand, when the iliofemoral ligament is tensioned, it prevents the trunk of the body from dropping backward. 
An iliofemoral ligament should be attached to the bending axis of hip joints winding around them as a human being. 
This mechanism is expected to work as a soft joint limit in the hip joints. We used the buckle mechanism (\figref{fig:buckle}) to attach ligament through the underside of the bending axis of hip joints (\figref{fig:application-hip-spiral}). 
This planar interskeletal structure uses the advantage of all merits described in section \ref{sec:musculoskeletal-structure} to wind around the hip joint and softly limit the range of joint angle.
}%
{%
股関節の腸骨大腿骨靭帯は大腿骨頭に巻き付くように螺旋状に繋がる靭帯である。主な役割は股関節の伸展の制限であり、股関節の屈曲時は緩むが、伸展時は大腿骨に巻き付いて緊張し体幹が後方へ倒れるのを防ぐ。筋骨格ヒューマノイドに適用する場合、人体と同様に股関節ジョイントの屈曲軸に巻きつけるように取り付けることで、股関節ジョイントのソフトな関節リミットとしての役割を果たすと考えられる。取り付け位置の余裕の関係から\figref{fig:application-hip-spiral}に示すように、人体と同じ巻きつき方向で屈曲軸の下側から巻き付くよう取り付けた。
}%

\subsubsection{Knee collateral ligaments}
\label{sec:application-knee-collateral-ligament}
\begin{figure}[htb]
  \centering
  \includegraphics[width=0.9\columnwidth]{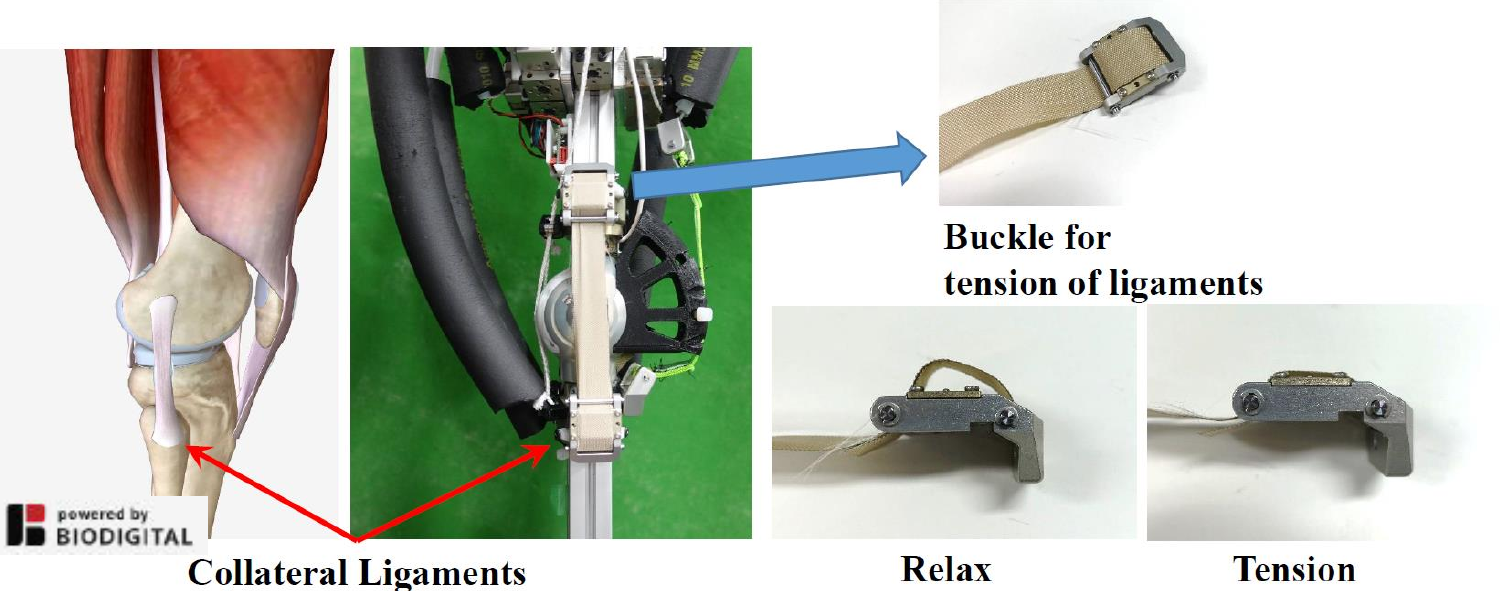}
  \caption{3D musculoskeletal model of the knee joint \cite{biodigital} and the application of the planar interskeletal structure to the collateral ligament. Collateral ligaments are fixed tensionally on the skeletal structure by the buckle method.}
  \label{fig:application-knee-collateral}
\end{figure}

\switchlanguage%
{%
    The collateral ligament in the knee joint softly restricts the movement of the knee joint in the bending position and stabilizes the knee joint in the extending position. 
We used the buckle mechanism to attach the collateral ligament to the skeletal structure. 
They were attached over the knee joint in the same positions as those in the human knee joint. 
The tension of the collateral ligament is adjusted in a soft bending position (\figref{fig:application-knee-collateral}). 
This planar interskeletal structure uses the advantage of all merits described in section \ref{sec:musculoskeletal-structure} to stably contact with skeletal sturctures and softly limit joint angles.
}%
{%
膝関節の側副靭帯は屈曲位ではやや弛緩しつつも関節の動きを制限し、伸展位では関節を内外の両側から拘束することで膝関節の安定性に貢献している。固定方法としては膝関節をまたいで人体と同様の位置にバックルを取り付けた。膝をやや屈曲した状態で靭帯の長さを変え、張力の調整を行う。
}%

\subsubsection{Patella ligaments}
\begin{figure}[htb]
  \centering
  \includegraphics[width=1.0\columnwidth]{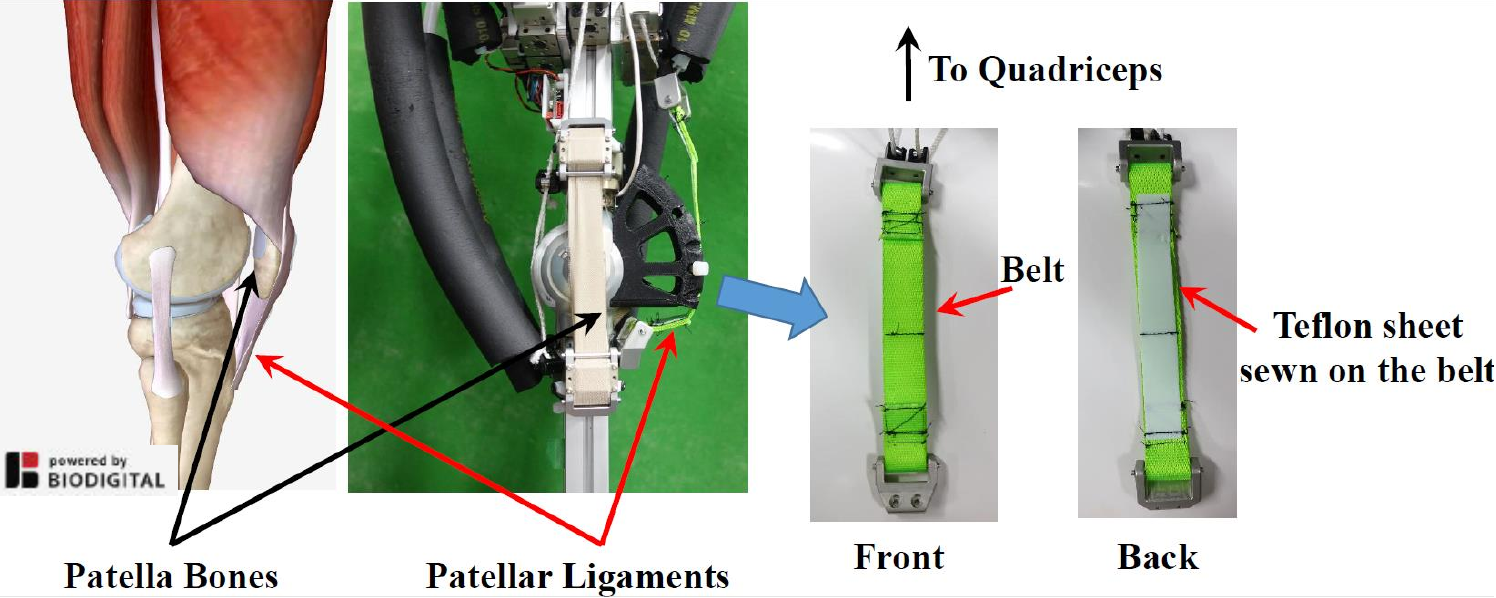}
  \caption{3D musculoskeletal model of the knee joint \cite{biodigital} and patella bone and patellar ligament applied on musculoskeletal humanoid as planar interskeletal structure. Teflon sheet is sewn on the backside of patellar ligament to reduce friction between the patellar ligament and the patella bone.}
  \label{fig:application-knee-patellar}
\end{figure}

\switchlanguage%
{%
The patella ligament connects a quadriceps femoris with a tibia. 
This ligament is not a muscle. 
However, in this study, we treat it as the actively drivable interskeletal structure, because its role is close to a muscle.

The patella ligament expands and contracts along the front side of the femur and patella bone. 
If there is no patella bone, the moment arm in the knee joint will decrease by ${20\%}$ \cite{jbjs:kaufer:patella}. 
Therefore to maintain the moment arm of the patella ligament and the quadriceps, the musculoskeletal humanoid needs the skeletal structure biologically equivalent to the patella bone. 
We used the skeletal component equivalent to the patella bone to maintain the moment arm.

To achieve smooth contact of the patella ligament with the patella bone, we used the belt attached to the muscles through the relay parts that fold back muscles. 
Besides, we sewed the Teflon sheet on the backside side of the belt, as shown in \figref{fig:application-knee-patellar}. 
This mechanism makes the patella ligament strong and decreases the friction between the patella ligament and the patella bone. 
This is one of the merits of planar interskeletal structure described in \ref{subsec:catching-prevention}. 
This planar interskeletal structure also uses the advantage of merits described in \ref{subsec:stable-contact} to stably transmit torque from muscle actuator to skeletal sturcture.
}%
{%
膝蓋骨靭帯は大腿四頭筋と脛骨を結ぶ靭帯であり厳密には筋ではないが、役割としては筋に近いため、能動的に駆動可能な骨格間構造として扱う。

膝蓋骨靭帯は膝蓋骨と共に大腿骨の前面に沿って滑って動き、大腿四頭筋のモーメントアーム増加に役立っている。人体に置いて膝蓋骨を取り除いた場合、膝のモーメントアームは20\%減少する\cite{jbjs:kaufer:patella}。したがって筋骨格ヒューマノイドに適用する場合、膝蓋骨靭帯および大腿四頭筋のモーメントアームの維持には、人体模倣の観点から生体力学的に同様の骨格構造が必要となる。そこで実装の際には、モーメントアームは膝蓋骨の代わりとなる骨格部品で確保することとした。

また骨格との滑らかな接触のため、設計解としてベルトとワイヤを経由パーツで結び、ベルトの裏側にテフロンシートを縫合することでベルトの強靭さとテフロンシートの摩擦特性を両者活かす方法をとった(\figref{fig:application-knee-patellar})。ここでは面状骨格間構造によって縫合という手法を取ることが出きたほか、機構への引っ掛かりを無くしている。
}%

\subsubsection{Gluteus maximus}
\begin{figure}[htb]
  \centering
  \includegraphics[width=0.9\columnwidth]{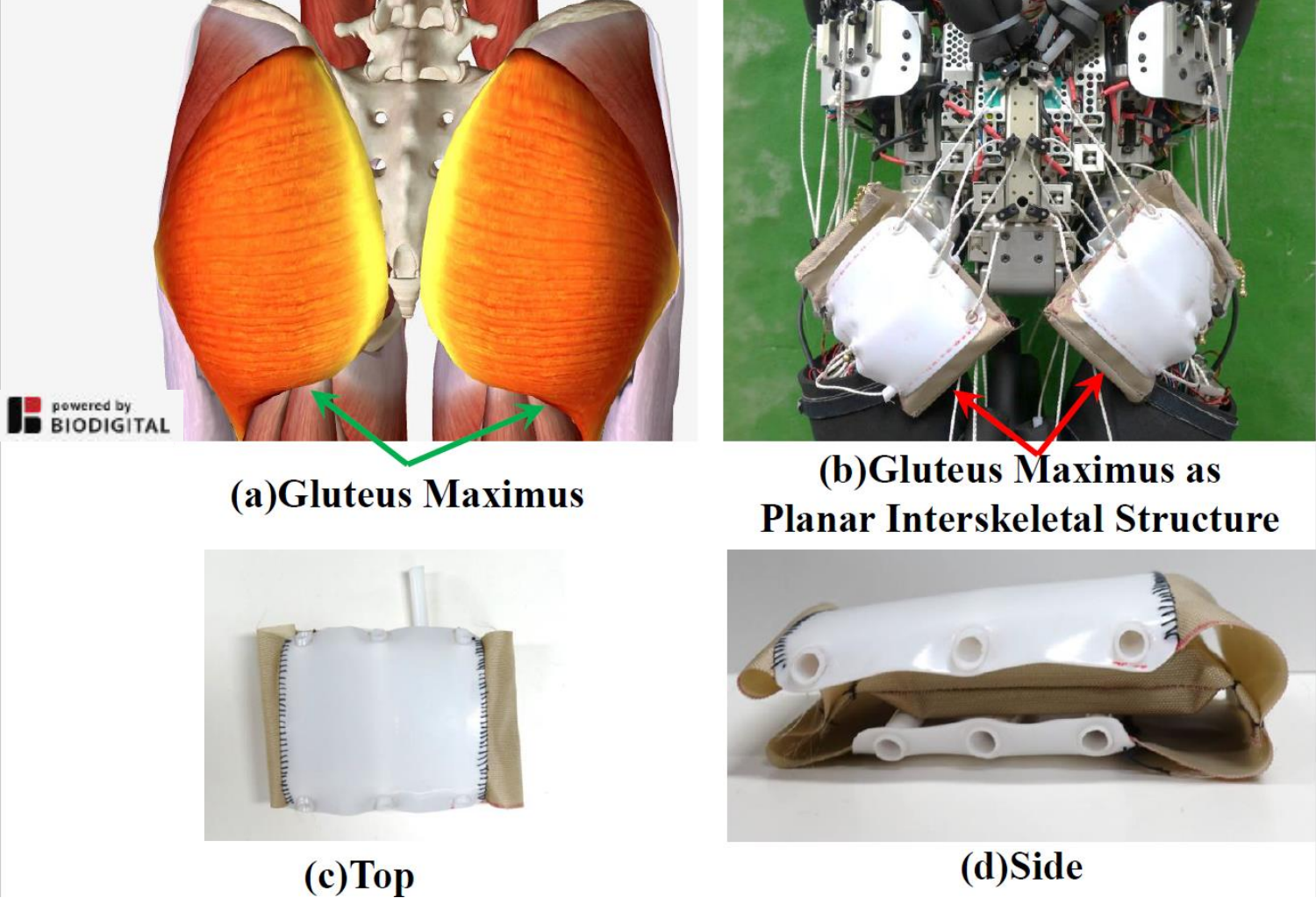}
  \vspace{-1.0ex}
  \caption{3D musculoskeletal model of the hip \cite{biodigital} and gluteus maximus applied as planar interskeletal structure on musculoskeletal humanoid. gluteus maximus have thick body so as to increase moment arm.}
  \label{fig:application-hip}
\end{figure}

\switchlanguage%
{%
    The gluteus maximus is the strongest muscle in the human body and is the crucial antigravity muscle \cite{book:kapandji:legs}. 
This muscle covers the hip joint, and the main role is extension and extorsion of the hip joint. 
Not only the surface contact by the planar interskeletal structure but also the thickness of the gluteus maximus is considered to play an essential role in maintaining the moment arm. 
In addition, we had to decrease the friction between the gluteus maximus and the hip joint to actuate the hip joint, because the gluteus maximus has a huge contact area with a skeletal structure and performs high torque sufficient to support the whole weight of the human body.

    In this study, we realized the thick gluteus maximus by using the pocket which contains polyethylene cushioning material sewn to the Teflon sheet, as shown in \figref{fig:application-hip}. 
To prevent friction between wires and skeletal structure, we implemented the Teflon tube to protect the wire path. 
Also, we achieved smooth contact of planar interskeletal structure and the hip joint by using the Teflon sheet sewn to the outside of the polyethylene pocket. 
This planar interskeletal structure uses advantage of the merit described in \ref{subsec:stable-contact} and \ref{subsec:catching-prevention}.
}%
{%
    大殿筋は身体で最も強大な筋であり、股関節を覆ってその伸展や外旋を担う、重要な抗重力筋である\cite{book:kapandji:legs}。筋骨格ヒューマノイドに適用する場合、面状骨格間構造による面接触だけでなく、人体における大殿筋が持つ厚みもまたモーメントアームの維持に寄与していると考えられる。また大殿筋がもつ広い面積による面接触に加えて、体重を支えるほどの強いトルクが生じることが想定されるため、関節の駆動のためには内力による股関節と大殿筋との摩擦を低減させる必要がある。
本研究では、実装の際にポリエチレン製の緩衝材を袋に詰めて縫合することで大殿筋に筋自体の厚みを持たせた。ワイヤと骨格・筋外装・環境・緩衝材との摩擦によって張力の減衰を防ぐため、ワイヤの経路にはテフロンチューブを用いている。
さらにテフロンシートを用いた筋外装を縫合することで面状骨格間構造と股関節との滑らかな接触を実現している。
最後に制作した大殿筋を筋骨格ヒューマノイドへ実装した外観を\figref{fig:application-hip}に示す。
}%

\section{Experiments} \label{sec:experiment}
\subsection{Basic Experiment of Planar Iliofemoral Ligament}
\begin{figure}[htb]
  \centering
  \includegraphics[width=0.9\columnwidth]{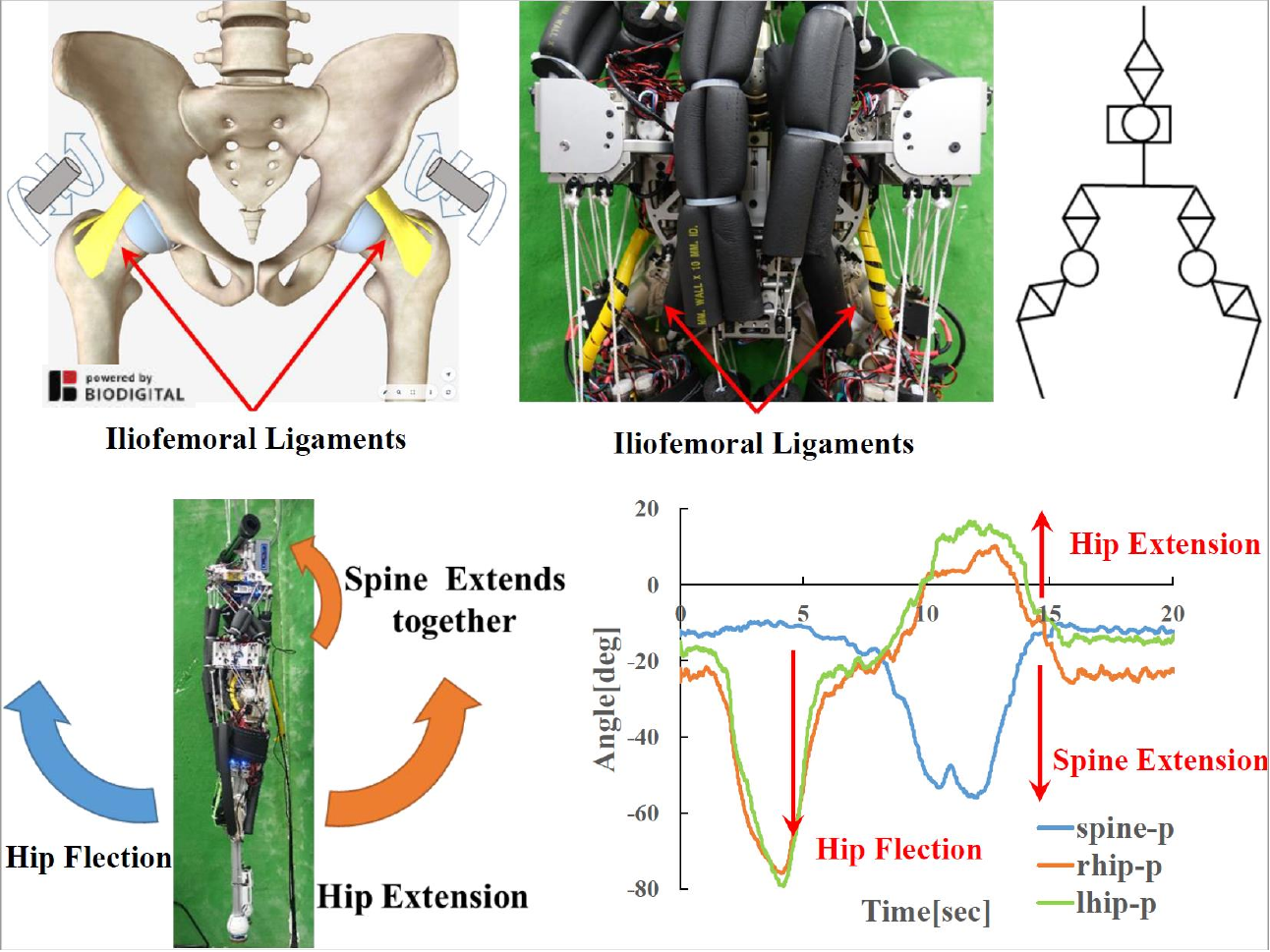}
  \vspace{-1.0ex}
  \caption{Overview of the hip and spine joint with planar iliofemoral ligaments. Relation between the pitch angle of the hip joint and the spine joint.}
  \label{fig:ovv-crotch}
\end{figure}

\switchlanguage%
{%
First, we will confirm that the iliofemoral ligament made of the elastic planar interskeletal structure can softly restrict the range of motion. 
In the human body, when a human moves the femur towards the backside of the body, tension is applied to the iliofemoral ligament and the iliofemoral ligament restricts the extension axis of the hip joint. 
This restriction makes the pelvis and the femur move together. 
We confirm that the hip joint of MusashiOlegs has the same function.

The external force is applied to the femur in both directions of extension and bending, on the condition of the trunk of the body being suspended. 
In order to evaluate the passive torque in the hip joint applied by the ligament, muscles around the pelvis are slackened. 
The result of this experiment is shown in \figref{fig:ovv-crotch}. 
In \figref{fig:ovv-crotch}, spine-p, rhip-p, lhip-p represent the joint angle of the spine pitch, rleg hip pitch and lleg hip pitch. When the hip joint is bent, the joint angle of the spine does not change. 
However, when the hip joint is extended, the passive torque by the iliofemoral ligament locks the hip joint and the joint angle of the spine changed.
}%
{%
面状骨格間構造による柔軟素材を用いた靭帯によって股関節における腸骨大腿靭帯を再現することで人体のように柔軟な可動域制限が実現されることを確認する。人体において大腿骨を体幹よりも後部に移動させる場合、靭帯の緊張によって股関節伸展軸が拘束され、骨盤ごと後傾する。MusashiOLegsの股関節においても同様の機能が実現されていることを確認する。
靭帯による関節受動トルクが生じていることを確認するため、骨盤周囲の筋はすべて弛緩させた。

体幹を吊り下げた状態で大腿骨に外部から力を加え、股関節を屈曲・伸展させる。この時の股関節屈曲伸展軸と脊柱の屈曲伸展軸の関節角度の様子を\figref{fig:hip-extension}右下の図に示す。股関節の屈曲時はフリージョイントである脊柱の関節角度は変化しないが、伸展時に腸骨大腿骨靭帯による関節受動トルクが生じて股関節がロックし、骨盤が後傾して脊柱の関節角度が変化していることが分かる。
}%

\subsection{Human Comparable Moving Function: Screw-Home Movement of Knee Joint}
\label{sec:human-comparable-SHM}
\switchlanguage%
{%
    In this experiment, we confirm that the screw-home movement in the knee joint is realized by the passive torque generated from the planar collateral ligament. 
The screw-home movement is the function of the planar collateral ligament. 
The ligament allows complex rotational movement in knee flexion configuration, but the ligament locks the range of angle and stabilizes the knee joint in knee extension configuration. 
The previous research realized screw-home movement by the only rigid mechanism \cite{IROS2013:asano:knee}. 
However, this mechanism cannot handle the impact force in the knee extension configuration and can break irreversibly. Not only the realization of a complex range of motion but also the soft restriction which can absorb impact force is important. 
We will show that the planar interskeletal structure implemented in \ref{sec:application-knee-collateral-ligament} can satisfy these characteristics.
}%
{%
人体の関節に見られる複雑な関節可動域の変化の例として膝関節の終末強制回旋機能が挙げられる。膝関節の終末強制回旋機能が面状側副靭帯による受動トルクによって実現されていること確認する。終末強制回旋機能とは、膝屈曲状態で回旋自由度を許して複雑な動作を可能にする一方、筋のモーメントアームが小さくなる膝伸展姿勢では靭帯と骨格形状、そして大腿四頭筋によって回旋角度を正面に収束させロックして関節を安定化させる機能である。筋骨格ヒューマノイドではリジッドな機構で実現した研究\cite{IROS2013:asano:knee}がなされているが、膝伸展時に衝撃がかかることによって軸が折れるなど致命的な破壊が起きうる。複雑な可動域の実現だけでなく人体のように腱や靭帯による衝撃吸収可能なソフトな拘束が重要であり、前節で実装した布による面状骨格間構造によってその両立が可能となる。
}%

\begin{figure}[htb]
 \centering
  \includegraphics[width=0.9\columnwidth]{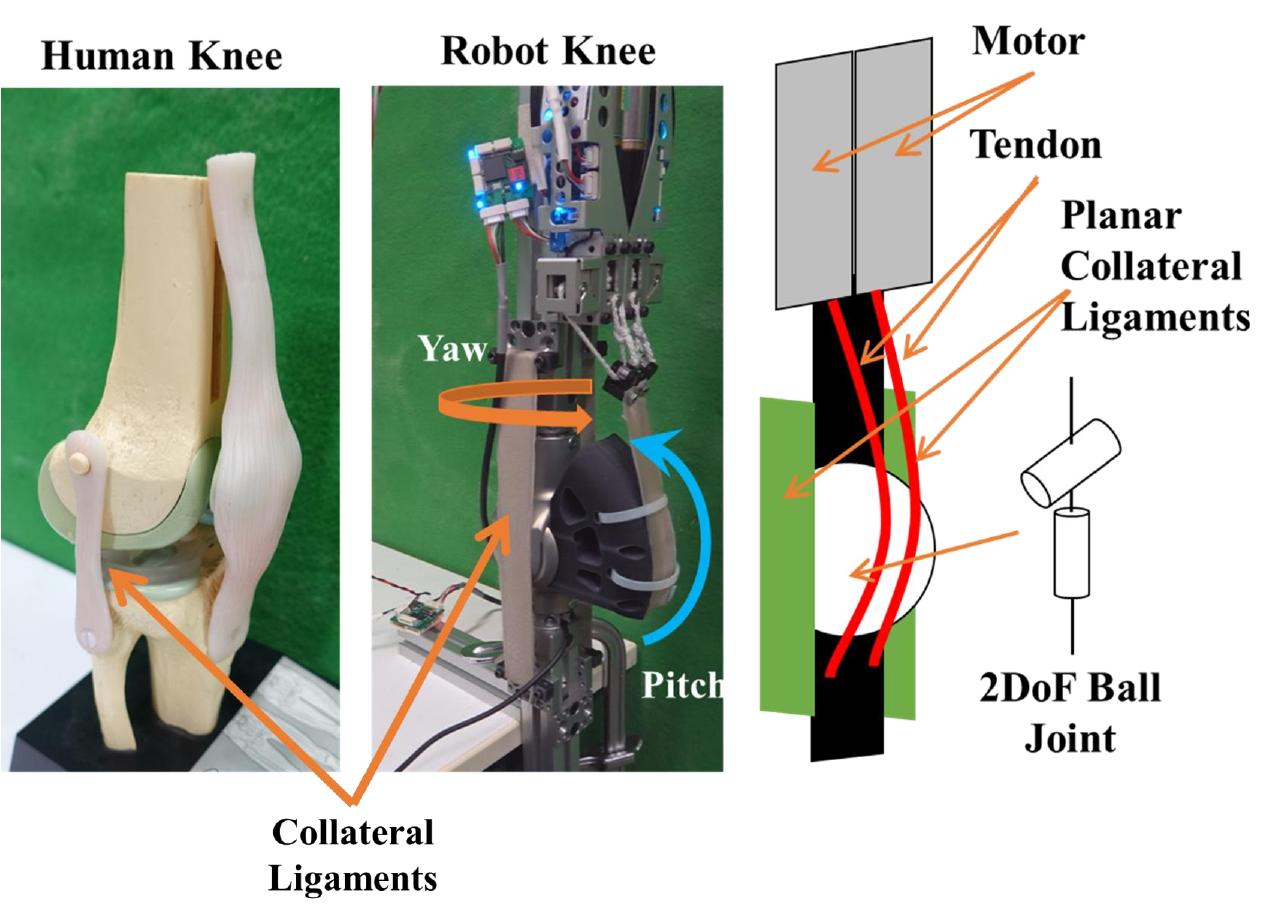}
  \vspace{-1.0ex}
  \caption{Overview of the knee joint with planar collateral ligaments and muscle unit.}
  \label{fig:ovv-knee}
\end{figure}

\begin{figure}[htb]
  \centering
  \includegraphics[width=1.0\columnwidth]{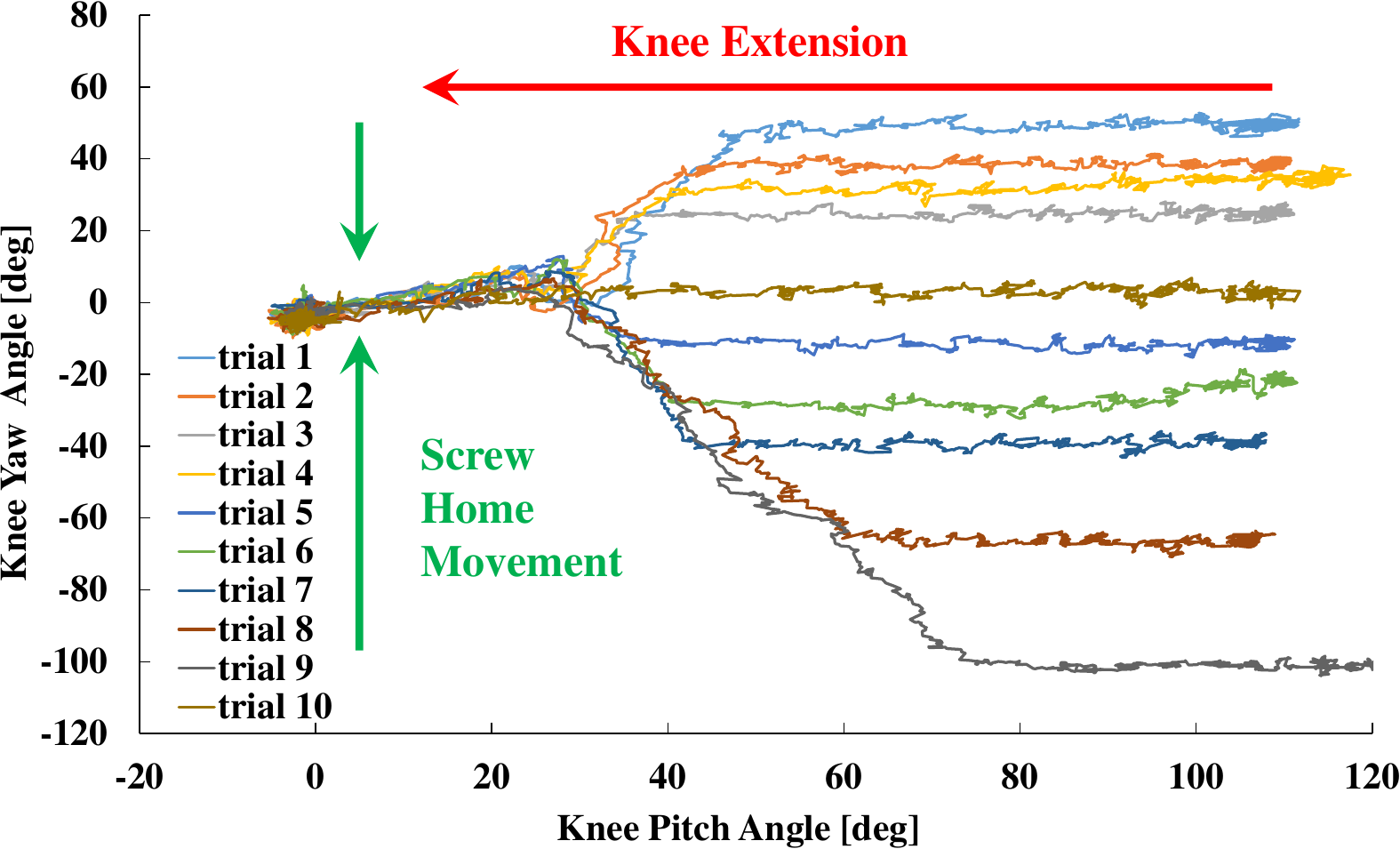}
  \vspace{-1.0ex}
  \caption{Relation between pitch angle and yaw angle of the knee joint during screw-home movement.}
  \label{fig:shm-graph}
\end{figure}

\begin{figure}[hbt]
 \centering
  \includegraphics[width=1.0\columnwidth]{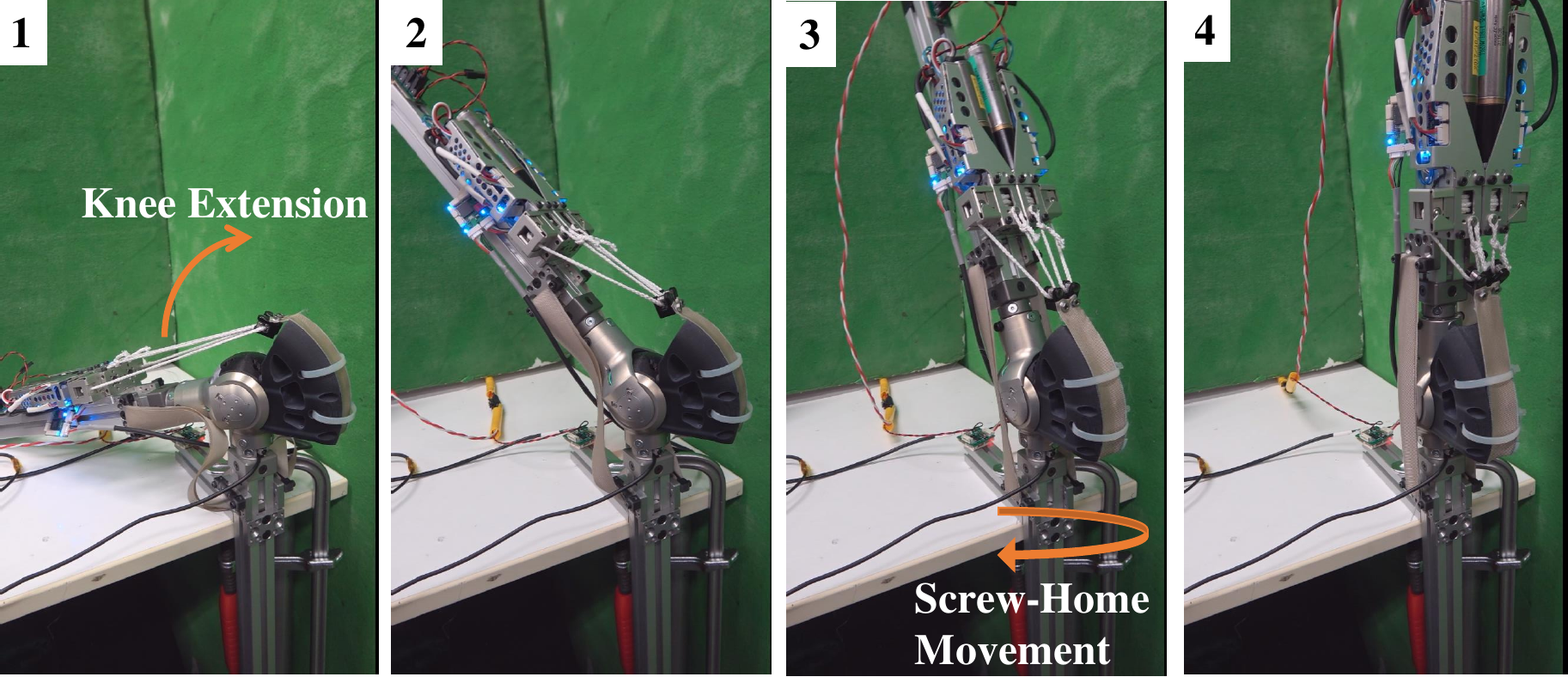}
  \vspace{-3.0ex}
  \caption{Screw-Home movement of the knee joint whose yaw angle is converged by collateral ligaments.}
  \label{fig:shm}
\end{figure}

\switchlanguage%
{%
    The schematic diagram of the knee joint attached to the collateral ligament and muscles is shown in \figref{fig:ovv-knee}. 
The collateral ligament is attached to the skeletal structure over the pseudo spherical joint which has 2Dof in yaw and pitch axis in \figref{fig:ovv-knee}. 
The quadriceps extend the knee joint. 
The collateral ligament should be attached to the skeletal structure of the thigh, because the screw-home movement is also the effect of the quadriceps. 
In this study, to evaluate only the effect of passive torque by the collateral ligament, the knee joint was fixed so that the quadriceps do not generate screw torque. 

    In these conditions, we set the knee joint bent and screwed randomly and extend knee joint using muscle actuators. 
\figref{fig:shm-graph} shows the result of 10 trials of extension of the knee joint from the randomly screwed configuration. 
This figure shows that the yaw joint angle of the knee joint converged to $0$ deg as the knee extends. 
As shown in \figref{fig:shm}, when the screwed knee joint extends, we observe that the screw-home movement by the passive torque of the collateral ligament locked the rotational axis of the knee joint as the knee joint extends.
}%
{%
\figref{fig:ovv-knee}に側副靭帯と筋を取り付けた膝関節の概要を示す。回旋と屈曲(\figref{fig:ovv-knee}のYawとPitch)方向の自由度を持つ擬似球関節を跨いで側副靭帯を骨格に取り付けた。大腿四頭筋によって膝関節を伸展させる。
終末強制回旋機能は大腿四頭筋によっても起きるため膝蓋骨靭帯は下腿の骨格に取り付けるべきだが、側副靭帯による関節受動トルクの影響のみを検証するため大腿の筋が回旋トルクを生じさせないように取り付けた。
膝を屈曲させた状態で回旋させて伸展させたところ、側副靭帯による関節受動トルクによって終末強制回旋機能が生じることが確認された(\figref{fig:shm})。膝関節の伸展に応じて正面方向へ関節がロックされている。
膝屈曲状態で回旋角度を適当にとり、10回にわたって膝を伸展させた際の膝の回旋角度の様子を\figref{fig:shm-graph}に示す。複数の姿勢から膝の伸展に応じて回旋が生じ、関節角度が収束していることが分かる。
    }%

\subsection{Performance of High Torque in a Wide Range of Motion}
\switchlanguage%
{%
    We conducted a squat motion experiment to show that the musculoskeletal humanoid legs with planar interskeletal structures can perform high torque in a wide range of motion. 
In this experiment, muscles must maintain a moment arm to support the body weight of humanoid legs in a wide range of motion. 
We send muscle length direction to each muscle actuators using learned body image \cite{IROS2018:kawaharazuka:OnlineSelfbodyImageLearning} which maps muscle length to joint angles.
}%
{%
面状骨格間構造による大殿筋と膝蓋骨靭帯によって、股関節及び膝関節における広い可動域においてモーメントアームが維持され十分なトルクが得られることを確認するため、これらの面状骨格間構造を取り付けた筋骨格ヒューマノイドMusashiOLegsにおいて屈伸動作を行った。
    }%
\begin{figure}[htb]
  \centering
  \includegraphics[width=1.0\columnwidth]{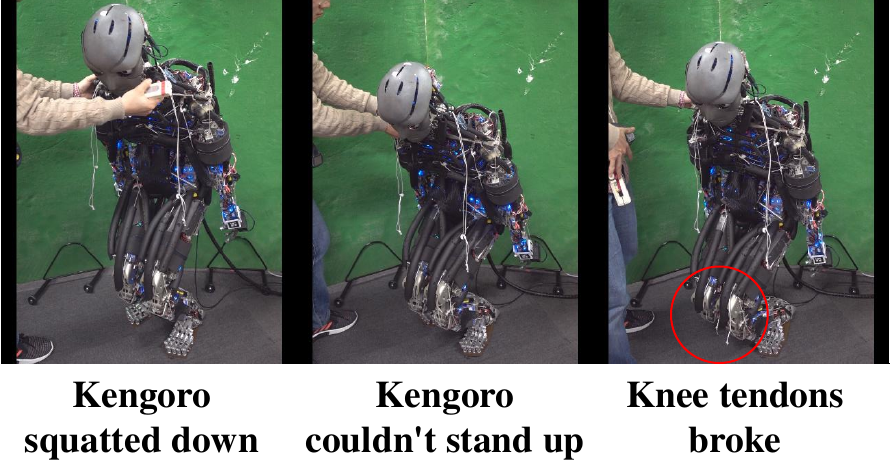}
  \vspace{-3.0ex}
  \caption{Squat motion using the previous musculoskeletal humanoid kengoro. The musculoskeletal humanoid without planar interskeletal structures could not perform squat motion.}
  \label{fig:kengoro-squat}
\end{figure}

\begin{figure}[htb]
  \centering
  \includegraphics[width=1.0\columnwidth]{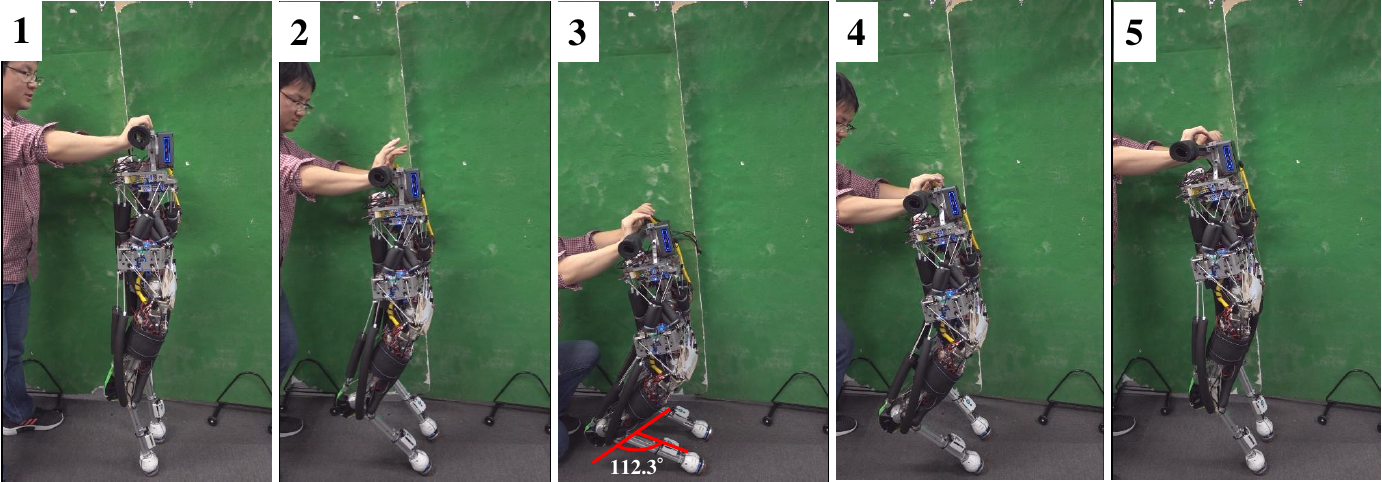}
  \vspace{-3.0ex}
  \caption{Demonstration of squat motion using MusashiOLegs. The musculoskeletal legs with the planar interskeletal structures succeeded to perform the deep squt motion.}
  \label{fig:musashiolegs-squat}
\end{figure}

\switchlanguage%
{%
    We show the result of the experiment in \figref{fig:kengoro-squat} and \figref{fig:musashiolegs-squat}. 
The musculoskeletal humanoid without the planar interskeletal structures \cite{Humanoids2016:asano:kengoro} could not perform the squat motion, because it could not apply sufficient torque to joints.
On the other hand, the musculoskeletal legs with the planar interskeletal structures achieved deep squat motion from the state \{${hip joint, knee joint: 64.7, 112.3}$\} ${[deg]}$.
}%
{%
動作の様子を\figref{fig:musashiolegs-squat}に示す。股関節は64.7度、膝関節は112.3度まで屈曲した状態から脚部を伸展させる深い屈伸動作を実現できた。
}%

\subsection{Pedal Switching Experiment as Environmental Contact Situation}
\begin{figure}[htb]
  \centering
  \includegraphics[width=1.0\columnwidth]{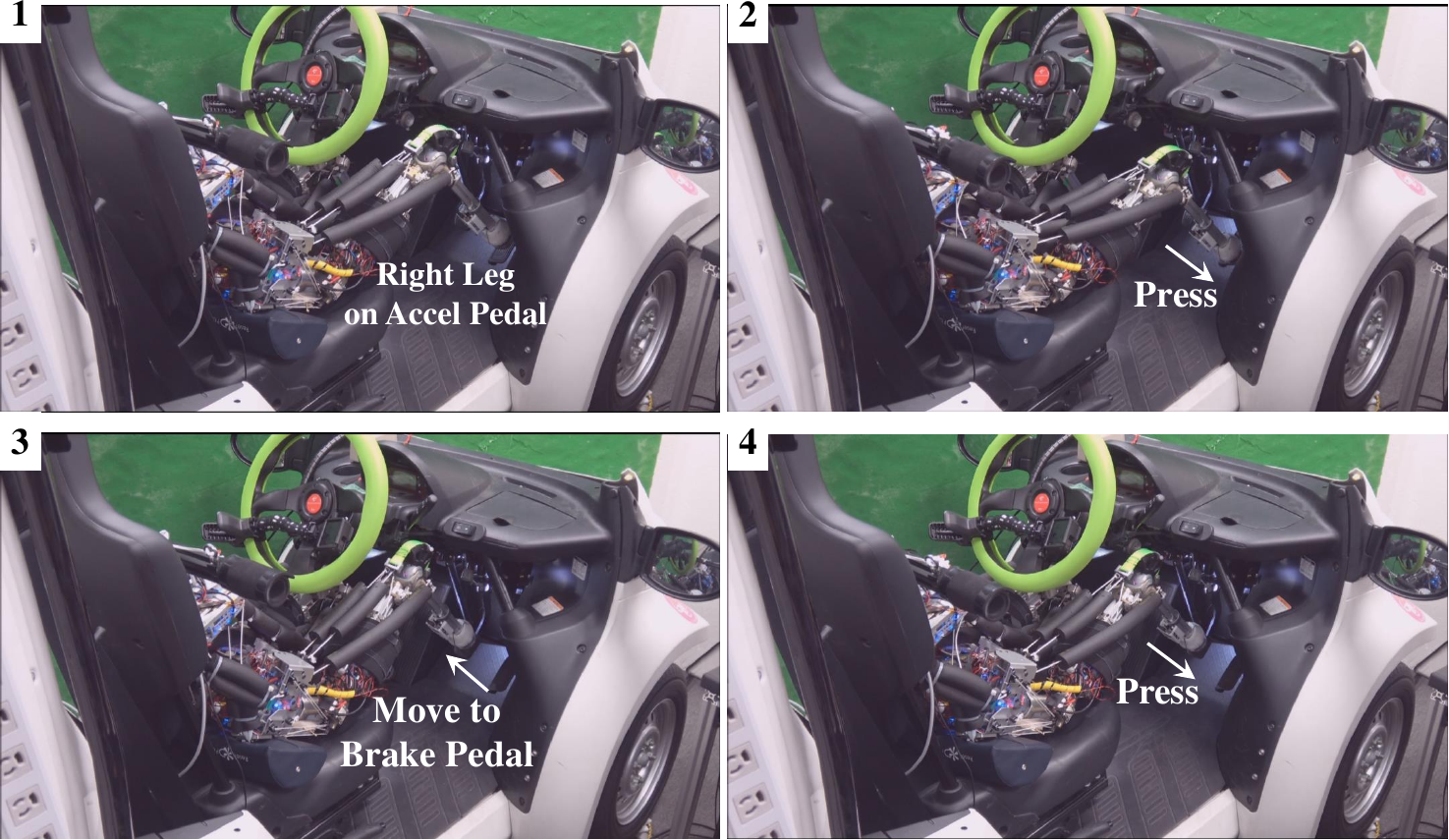}
  \vspace{-3.0ex}
  \caption{MusashiOLegs performing the pedal switching from the accelerator pedal to the brake pedal on the driving sheet.}
  \label{fig:out-driver-sheet}
\end{figure}

\switchlanguage%
{%
Finally, we conducted the pedal switching experiment as environmental contact situation. 
In the pedal switching experiment, muscles must maintain the moment arm to transmit sufficient torque to each joint in tight postures. 
Also, the musculoskeletal legs must perform the task under friction with the car seat. 
In this experiment, the musculoskeletal legs is required to, first press the accel pedal, second put its foot away from the accel pedal, third move its foot to the above position of the brake pedal and finally press the brake pedal. 
The car used in this experiment is B.COM Delivery of extremely small EV COMS series. 
We send muscle-length direction to each muscle actuator using learned body image \cite{IROS2018:kawaharazuka:OnlineSelfbodyImageLearning} which maps muscle length to joint angles.
}%
{%
本節では椅子に座った状態での動作として行った運転時のペダル踏み変え動作について述べる。自動車にはトヨタ車体製の超小型EV「コムス」シリーズのB-COMデリバリーを使用した。本動作の流れを以下に説明する。MusashiOLegsを車の座席に座らせ、右脚部によるアクセルペダルとブレーキペダルの踏み変え動作を行う。
従来の筋骨格ヒューマノイドでは座席との摩擦などで脚部を動かすことが出来ず、両足でそれぞれのペダルを踏んでいた。
}%

\begin{figure}[htb]
  \centering
  \includegraphics[width=1.0\columnwidth]{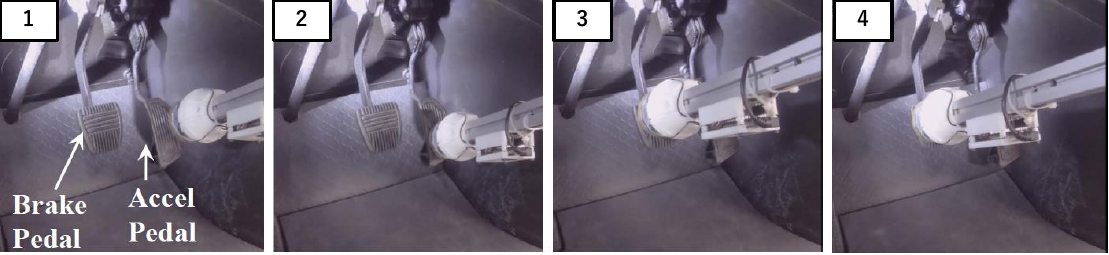}
  \vspace{-3.0ex}
  \caption{Pedal switching motion of MusashiOLegs on the driving sheet from lower viewpoint.}
  \label{fig:in-driver-sheet}
\end{figure}

\begin{figure}[htb]
  \centering
  \includegraphics[width=0.9\columnwidth]{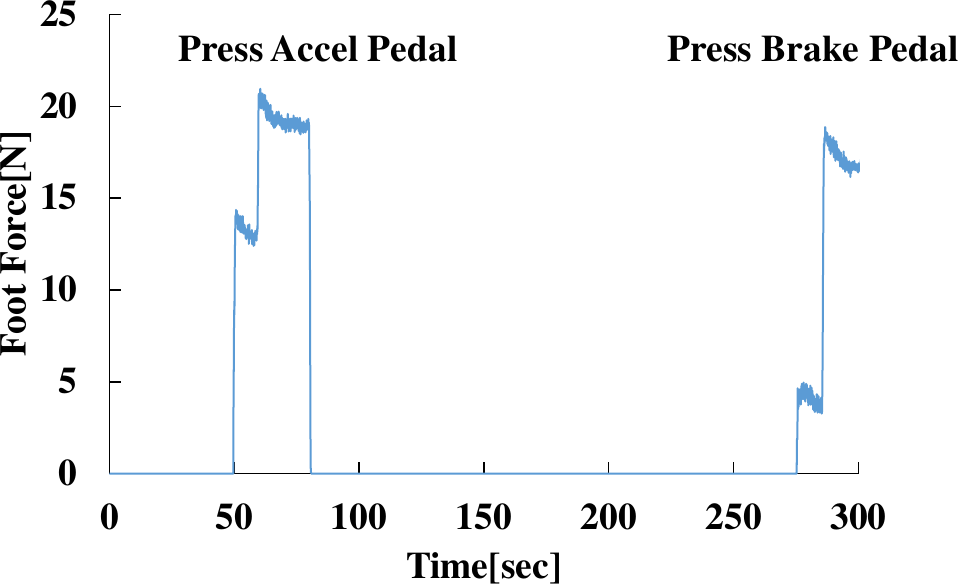}
  \vspace{-1.0ex}
  \caption{Right foot force during pedal switching motion.}
  \label{fig:pedal-force-sensor}
\end{figure}

\switchlanguage%
{%
The experimental appearance is shown in \figref{fig:out-driver-sheet}. 
In addition, \figref{fig:in-driver-sheet} shows the appearance from the view above feet and \figref{fig:pedal-force-sensor} shows the foot force sensor data during the experiment. 
In previous research, the musculoskeletal humanoid pressed pedals using one leg for each pedal; the left leg for the brake pedal, and the right leg for the accel pedal \cite{IROS2019:kawaharazuka:pedal-control}. 
This was because that the musculoskeletal humanoids could not produce enough torque to raise each leg in a seated situation. 
On the other hand, the musculoskeletal humanoid legs with planar interskeletal structures achieves pedal switching by single-leg like a human being. 
It can produce enough torque to raise the leg in the seated situation desptie friction between the car seat and the body.
}%
{%
本実験の動作の様子について、運転席内部を\figref{fig:in-driver-sheet}に、外から見た運転席を\figref{fig:out-driver-sheet}に、足先の力センサの値のグラフを\figref{fig:pedal-force-sensor}示す。狭隘な運転席内部で広い関節可動域を生かして脚を動かしペダル踏み変え動作を実現している。この際、柔軟なドライビングシートと身体をなじませつつ、座面との摩擦のなかで脚部を動かすだけのトルクを発揮することが出来ている。また筋と骨格との内力もまた、テフロンによる筋膜によって抑えられている。
}%


\section{CONCLUSION} \label{sec:conclusion}
\switchlanguage%
{%
In this study, we discussed the role of interskeletal structures that cover joints and tract different skeletal structures and showed the importance of planar interskeletal structures. 
We showed that the planar interskeletal structure has three key features; stable contact with the surface of the skeletal structure, prevention of being caught in folds of structure and the tolerance of the tension in the shear direction. 
About the tolerance of the tension in the shear direction, we showed planar structure has better resistance compared to liner structures, using simulation. Then we designed the musculoskeletal legs with planar interskeletal structures. 
We applied passive planar interskeletal structures to iliofemoral ligaments and knee collateral ligaments in order to softly restrict joint angles in the angle limit.
Active planar interskeletal structures were applied to patella ligaments and gluteus maximus in order to maintain the moment arm of muscle and stably transmit torque to skeletal structures.
Also, we conducted experiments that require human comparable moving function and verified the effectiveness of the planar interskeletal structures. 
In future works, we would like to investigate control systems utilizing the high joint stiffness and stable torque performance.
Also, we would like to realize the learning based controller which can control the behavior of planar interskeletal structures.
}%
{%
本研究では人体模倣を規範とする筋骨格ヒューマノイドにおいて、関節を跨いで骨格間を牽引する筋や靭帯などの骨格間構造に着目し、関節に安定してトルクを及ぼすために人体に見られる面状構造の重要性について述べた。筋骨格ヒューマノイドを開発し、面状骨格間構造の適用を含めて議論し、実際のロボットにおいて人体と同様の動作機能が実現されていることを確認した。これによって筋骨格ヒューマノイドにおける面状骨格間構造の重要性を示すことが出来た。

今後は面状骨格間構造の特性を含んだ動作学習を行い、安定したトルク発揮による高関節剛性を活用した動作へ展開させていきたい。
}%

{
  \bibliographystyle{IEEEtran}
  \bibliography{main}
}

\end{document}